\pdfoutput=1

\documentclass[11pt]{article}

\usepackage[]{acl}

\usepackage{times}
\usepackage{latexsym}

\usepackage[T1]{fontenc}

\usepackage[utf8]{inputenc}

\usepackage{microtype}

\usepackage{url}
\usepackage{times}
\usepackage{latexsym}

\usepackage{multicol}
\usepackage{multirow}
\usepackage{graphicx}
\usepackage{verbatim}
\usepackage{makecell}
\usepackage{braket}
\usepackage{hhline}
\usepackage{amsmath,bm,amssymb}
\usepackage[linesnumbered,ruled]{algorithm2e}
\usepackage{xcolor}
\definecolor{Green}{rgb}{0.0, 0.5, 0.0}
\definecolor{Amethyst}{rgb}{0.6, 0.4, 0.8}
\DeclareSymbolFont{extraup}{U}{zavm}{m}{n}
\DeclareMathSymbol{\varheart}{\mathalpha}{extraup}{86}
\DeclareMathSymbol{\vardiamond}{\mathalpha}{extraup}{87}

\title{What are the Desired Characteristics of Calibration Sets? \\ Identifying Correlates on Long Form Scientific Summarization }

\author{
  Griffin Adams$^{\spadesuit,\clubsuit}\thanks{\hspace{0.2em}  Work started during internship with Microsoft Research.}$ \\ griffin.adams@columbia.edu \And
  Bichlien H Nguyen$^{\diamondsuit}$ \\ bnguy@microsoft.com \\ \And Jake Smith$^{\diamondsuit}$ \\ jakesmith@microsoft.com \\ \AND
  Yingce Xia$^{\diamondsuit}$ \\ yingce.xia@microsoft.com \And
  Shufang Xie$^{\diamondsuit}$ \\ shufxi@microsoft.com \And
  Anna Ostropolets$^{\clubsuit}$ \\ anna.ostropolets@columbia.edu \\ \AND
  Budhaditya Deb$^{\diamondsuit}$ \\ budha.deb@microsoft.com \And
Yuan-Jyue Chen$^{\diamondsuit}$ \\ yuanjc@microsoft.com \And
  Tristan Naumann$^{\diamondsuit}$ \\ tristan@microsoft.com \\\AND No\'emie Elhadad$^{\spadesuit,\clubsuit}$ \\ noemie.elhadad@columbia.edu \AND
 Microsoft Research$^{\diamondsuit}$ \quad  Columbia University: Computer Science$^{\spadesuit}$, Biomedical Informatics$^{\clubsuit}$
}
\begin{document}
\maketitle
\begin{abstract}

Summarization models often generate text that is poorly calibrated to quality metrics because they are trained to maximize the likelihood of a single reference (MLE). To address this, recent work has added a calibration step, which exposes a model to its own ranked outputs to improve relevance or, in a separate line of work, contrasts positive and negative sets to improve faithfulness. While effective, much of this work has focused on \emph{how} to generate and optimize these sets. Less is known about \emph{why} one setup is more effective than another. In this work, we uncover the underlying characteristics of effective sets. For each training instance, we form a large, diverse pool of candidates and systematically vary the subsets used for calibration fine-tuning. Each selection strategy targets distinct aspects of the sets, such as lexical diversity or the size of the gap between positive and negatives. On three diverse scientific long-form summarization datasets (spanning biomedical, clinical, and chemical domains), we find, among others, that faithfulness calibration is optimal when the negative sets are extractive and more likely to be generated, whereas for relevance calibration, the metric margin between candidates should be maximized and surprise--the disagreement between model and metric defined candidate rankings--minimized. Code to create, select, and optimize calibration sets is available at \url{https://github.com/griff4692/calibrating-summaries}.
\end{abstract}

\section{Introduction}

Traditionally, summarization models have been trained to maximize the likelihood of gold-standard references. This training paradigm introduces an exposure bias because, during training, the model is not exposed to the metrics on which it is evaluated. Without being able to calibrate its own predictions with metrics, models are prone to produce summaries with irrelevant or repetitive content \citep{zhao2022calibrating}, or misrepresent the claims in the source text \citep{cao2018faithful, maynez-etal-2020-faithfulness}.

Calibration offers a flexible and effective set of methods to remedy this exposure bias by explicitly instructing a model to distinguish between high and low quality summaries. By varying how candidate sets are constructed and optimized, an extra calibration step can unlock large gains in relevance (via ROUGE \citep{simcls, liu-etal-2022-brio}) or improve the faithfulness of summaries to the source \citep{nan-etal-2021-improving, cliff}.

Yet, much of this work has addressed \emph{how}---how to generate candidates \citep{cliff} and how to define effective calibration objectives \citep{nan-etal-2021-improving, zhao2022calibrating}. Work has largely been separated into relevance and faithfulness calibration, with less study of the interaction between the two. Relevance, often measured with ROUGE, captures the content overlap with a human-written reference, whereas faithfulness is typically reference-free, and captures the fidelity of a summary to the source text(s). In this paper, we examine both faithfulness and relevance as the target metrics for calibration and seek to uncover the underlying characteristics of effective calibration sets for each separately, as well as analyze the interactions between them. To accomplish this, we implement a diverse set of existing methods for constructing candidate and corrupted summaries and combine them to form a large candidate pool. From this pool, we implement different filtering strategies for set selection, which target specific characteristics, such as the metric margin between negatives and positives, diversity, and the model likelihood of generating each candidate in the set.

We run experiments that vary only in the training data selected for candidate sets. For each experiment, we extract a wide range of relevant statistics (e.g., diversity, length) on the candidate sets and show the relationship between these set statistics and downstream performance. To guide future research, we analyze the plots to provide insights into, and rationale for, optimal set construction.

Additionally, a large portion of research has focused on summarization of single-document news articles \citep{gehrmann2022repairing, mckeown-keynote}. We seek to broaden and pressure test recent advances in contrastive fine-tuning by experimenting on three long-form, scientific, highly specialized corpora in which metrics, e.g. faithfulness, are non-trivial to define, capture, and categorize. Also, long-form summarization is appealing for our calibration experiments given that memory is constrained. Even with training tricks, such as gradient accumulation and half precision, only a small handful of candidates per example (4 in our experiments\footnote{Each experiment was run on a relatively large card with 40GB of GPU memory (the NVIDIA A100).}) fit in memory. This makes the selection step more important compared to shorter tasks.



The primary contributions of this work are to: \textbf{(1)} benchmark calibration models on three scientific long-form datasets, including a new, chemistry-focused corpus, for which we collect fine-grained faithfulness annotations and relevance rankings from experts; \textbf{(2)} conduct extensive experiments to better understand the underlying characteristics and dynamics of effective calibration tuning sets. We release easily extensible code for forming and optimizing calibration sets in the scientific domain.


\section{Related Work}

Typically, when summarization models are calibrated to quality metrics, this refers to contrastive learning to improve faithfulness. Contrastive learning for faithfulness has been applied to fine-tuning \citep{nan-etal-2021-improving, tang-etal-2022-confit, cliff}, post-hoc editing \citep{cao-etal-2020-factual, zhu-etal-2021-enhancing}, re-ranking \citep{chen-etal-2021-improving}, and evaluation \citep{factcc, wu-etal-2020-unsupervised, deng-etal-2021-compression}. This line of research has largely focused on the methods used to generate synthetic errors for negative contrast sets: i.e., by directly mimicking errors observed during human evaluation \citep{tang-etal-2022-confit}, entity swapping \citep{cliff}, language model infilling \citep{cliff}, or using unfaithful system outputs \citep{nan-etal-2021-improving}. Orthogonal to our work, \citet{cliff} assess the relative efficacy of a diverse set of corruption methods when used for contrastive fine-tuning for faithfulness.


For relevance calibration, models are typically calibrated to the ROUGE scores of their own outputs after an initial fine-tuning step \citep{liu-liu-2021-simcls, liu-etal-2022-brio}. \citet{zhao2022calibrating} extend the work of \citet{liu-etal-2022-brio} and run a broad sweep of loss functions and candidate generation methods for two-step relevance calibration while establishing state of the art performance (ROUGE) across single document corpora. As opposed to contrasting positives and negatives in a latent space, these models are instructed to calibrate decoder likelihoods to ROUGE or BERTScore-defined rankings.


Our work is distinct along three key dimensions: \textbf{(1)} we consider long-document scientific summarization, rather than single-document; \textbf{(2)} we consider both faithfulness and relevance calibration and analyze the interactions between the two, often competing, quality objectives; \textbf{(3)} we uncover relationships between key set statistics and downstream performance by systematically varying how calibration sets are formed from candidate pools.

\section{Datasets}

\begin{table}[t]
\centering
\small
\begin{tabular}{l|c|c|c}
\textbf{Statistic} & \textbf{Clinical} & \textbf{Chemical} & \textbf{Bio.} \\ \hline
Train Size & 41,705 & 115,956 & 119,924 \\ 
Validation Size & 940 & 1,000 & 6,633 \\
Test Size & 1,861 & 2,000 & 6,658 \\ \hline
Source Tokens & 8,175 & 5,364 & 3,092 \\
Reference Tokens & 416 & 216 & 205 \\  \hline
Extractive Coverage & 0.66 & 0.90 & 0.88 \\
Extractive Density & 1.97 & 3.53 & 5.87 \\
\hline
\end{tabular}
\caption{Statistics for long-form scientific summarization datasets. The biomedical dataset is from \citet{cohan-etal-2018-discourse}, the recipe to recreate the clinical from \citet{adams2022learning}, and the chemical from this work. } \label{tab:dataset-stats}

\end{table}

Dataset statistics are shown in Table~\ref{tab:dataset-stats}.

\paragraph{Clinical.} We use the long-form hospital course summarization dataset from \citet{adams2022learning}. Refer to Appendix \ref{app:clinical-dataset} for details on this dataset.

\paragraph{Chemical.} We introduce a dataset with a pure chemistry focus by compiling a list of chemistry academic journals with Open-Access articles. For each journal, we downloaded full-text article PDFs from the Open-Access portion of the journal using available APIs, or scraping this content using \href{https://www.selenium.dev/documentation/webdriver/}{Selenium Chrome WebDriver}. 
Each PDF was processed with Grobid \citep{lopez2009grobid} via a \href{https://pypi.org/project/grobid-client-python/}{client} to extract free-text paragraphs with sections. The inputs for the summarization models are section headers and associated paragraphs for all sections from Introduction through Conclusion, excluding references, tables, and image captions. The abstract is treated as the reference. While other scientific summarization datasets exist \citep{lu-etal-2020-multi-xscience, gupta-etal-2021-sumpubmed, deyoung-etal-2021-ms}, ours is the first to exclusively contain chemistry-related papers. \begin{table}[h]
\centering
\small
\begin{tabular}{l|c}
\textbf{Source} & \textbf{\# Articles}  \\ \hline
Beilstein & 1,829 \\
Chem Cell & 546 \\
ChemRxiv & 12,231 \\
Chemistry Open & 398 \\
Nature Communications Chemistry & 572 \\
PubMed Author Manuscript & 57,680 \\
PubMed Open Access & 29,540 \\
Royal Society of Chemistry (RSC) & 9,334 \\
Scientific Reports - Nature & 6,826 \\
\hline
\end{tabular}
\caption{Journals accessed for Chemical papers.} \label{tab:chemistry-sources}
\end{table}

Table \ref{tab:chemistry-sources} shows the journals from which Open Access articles were sourced, as well as the number of papers processed. For all journals, we filtered for papers with the provided topic of Chemistry when papers from other disciplines were also available (e.g. PubMed). We randomly split the aggregated dataset into train-validation-test splits.

The dataset is available for download on the HuggingFace Datasets Hub under \href{https://huggingface.co/datasets/griffin/ChemSum}{griffin/ChemSum}.

\paragraph{Biomedical.} We use the PubMed abstract generation dataset \citep{cohan-etal-2018-discourse}, which pairs automatically extracted abstracts with full-text articles from the PubMed Open-Access Subset.

\section{Calibration Pipeline}

At a high-level, we fine-tune (\texttt{FT}) language models with standard maximum likelihood estimation (\texttt{MLE}) on each summarization corpus, and then \textit{calibration}-tune (\texttt{CT}) on a combined objective, which adds a calibration loss (\texttt{CA}) to the MLE loss:


\begin{align}
\label{eq:pipeline}
\begin{split}
&\mathcal{L}_{FT} = \mathcal{L}_{MLE} \\
&\mathcal{L}_{CT} = \lambda_{MLE} * \mathcal{L}_{MLE} + \lambda_{CA} * \mathcal{L}_{CA}
\end{split}
\end{align}

\noindent $\lambda_{MLE}, \lambda_{CA}$ are scalars controlling the relative weight of objective. For $\mathcal{L}_{CT}$, $\mathcal{L}_{MLE}$ acts as a regularizer, as in \citet{liu-etal-2022-brio, zhao2022calibrating}.



We describe the setup (objective, metrics, and candidate generation methods) for Relevance Calibration (\S \ref{sec:relevance-setup}) and Faithful Calibration (\S \ref{sec:faithful-setup}, before jointly discussing statistics on each setup (\S \ref{sec:calibration-analysis}).


\subsection{Relevance Calibration} \label{sec:relevance-setup}

As in \citep{liu-etal-2022-brio, zhao2022calibrating}, we calibrate for relevance by learning to rank model-generated summaries (post-\texttt{FT}, pre-\texttt{CT} weights).

\paragraph{Objective.} Specifically, a set of model-generated summaries $\bm{\hat{S}}$ is ranked: $q(\hat{S}_i; S) \geq q(\hat{S}_j; S)$, $\forall i, j \in |\bm{\hat{S}}|, i < j$, where $S$ is the reference and $q$ represents $Rel_{Agg}$ (defined below). A score function $f$ is applied to each candidate and calibrated to the metric ranking via a pairwise margin:


\begin{align}
\small
\label{eq:pmr}
\begin{split}
max(0, f(D, \hat{S}_j) - f(D, \hat{S}_i) + (j - i) * \lambda_{margin}) \\
\forall i, j \in |\bm{\hat{S}}|, i < j
\end{split}
\end{align}


$f$ represents for the length normalized log likelihood of generating a summary \citep{liu-etal-2022-brio}.




\paragraph{Rank Metric.} To define a gold-standard ordering, we aggregate 3 relevance metrics which are normalized to be zero after fine-tuning \texttt{FT}. $Rel_{Agg}$, a combination of ROUGE 1/2 F-1 \citep{lin2004rouge} and \textbf{BERTScore-Ref} \citep{zhang2019BERTScore}, represents the standard deviation change in the aggregated metric from \texttt{FT}. Full details are in Appendix \ref{app:metrics}.

\paragraph{Candidates.} We fine-tune (\texttt{FT}) two state of the art long-document language models: LongT5 \citep{longt5} and PRIMERA \citep{primera}, on each corpus before decoding 10 candidates with diverse beam search \citep{vijayakumar2016diverse} with diversity penalty of $1.0$, as in \citet{liu-etal-2022-brio}.


\begin{table*}[t]
\centering
\small
\begin{tabular}{cl|c|c|c|c|c|c}
& \textbf{Method} & \textbf{\textcolor{red}{$-$}} & \textbf{\textcolor{Green}{$+$}} & \textbf{Source} & \textbf{Ref.} & \textbf{External Components} & \textbf{Models Used} \\
\hline
\multirow{2}{*}{\texttt{\makecell{Relevance \\ Calibration}}} & Diverse Beam & \textcolor{red}{\checkmark} & \textcolor{Green}{\checkmark} & \checkmark & & Summarization Model & PRIMERA \\
& Diverse Beam & \textcolor{red}{\checkmark} & \textcolor{Green}{\checkmark} & \checkmark & & Summarization Model & LongT5 \\ \hline
\multirow{4}{*}{\texttt{\makecell{Faithful \\ Calibration}}} & Mask-And-Fill & \textcolor{red}{\checkmark} &  & & \checkmark & Constituency Parser, PLM & Stanza, SciFive \\
& Entity Swap & \textcolor{red}{\checkmark} & & & \checkmark & Entity, Number Extractors & BERN2, Quantulum \\
& Paraphrase & & \textcolor{Green}{\checkmark} & & \checkmark & Paraphrase Generator & GPT-3 + Curated Prompt \\
& Reference & & \textcolor{Green}{\checkmark} & & \checkmark & N/A & N/A \\ \hline
\end{tabular}
\caption{Methods to create \textcolor{red}{negative} and \textcolor{Green}{positive} candidates in support of relevance and faithfulness calibration, respectively. For each candidate generation method, we include whether it is used as a positive or negative example (both in the case of relevance ranking), what inputs it requires (the source document and/or the reference (ref.)), as well as the external components needed and, finally, the specific models used for the experiments in this paper.} \label{tab:method-overview}
\end{table*}

\subsection{Faithfulness Calibration} \label{sec:faithful-setup}

\paragraph{Objective.} As in \citet{gunel2020supervised, khosla2020supervised, cliff}, we use contrastive learning to minimize the latent distance between pairs of positive summaries vis-a-vis negative ones:


\begin{equation}
\label{eq:contrast}
\small
- \frac{1}{{|\bm{\hat{S}^P}| \choose 2 }} \sum_{ \hat{S}_i, \hat{S}_j \in \bm{\hat{S}^P}  }{ log \frac{exp(sim(h_i, h_j) / \tau )}{ \sum_{\hat{S}_k \in \bm{\hat{S}^N}}{exp(sim(h_i, h_k) / \tau)} } }
\end{equation}

where $\tau$ is a temperature parameter. It pushes positive summaries closer to each in latent space ($h_i$ and $h_j$) and further away from negatives ($h_k$). We follow \citet{cliff} and use cosine similarity as $sim$ and treat $h$ as the mean-pooled decoder states, followed by a linear projection.

\paragraph{Faithfulness Metric.} Similar to $Rel_{Agg}$, we compute $Faith_{Agg}$ as an aggregation of normalized metrics. We combine \textbf{BARTScore} \citep{yuan2021BARTScore}, \textbf{BERTScore-Src} (vis-a-vis source), and a new metric \textbf{FactScore}, which is based on a scientific fact detection model (MultiVERS \citep{wadden-etal-2022-multivers}). Full details are in Appendix \ref{app:metrics}.

\paragraph{Negative Methods.} We use an in-domain LM (SciFive) to \textbf{Mask-And-Fill} hallucinations, as well as perform \textbf{Entity Swaps} of scientific concepts and numbers which separately target \texttt{intrinsic} and \texttt{extrinsic} hallucinations \citep{maynez-etal-2020-faithfulness}. Please refer to Appendix \ref{app:neg-methods} for more details. \linebreak


\paragraph{Positive Methods.} We pool together the \textbf{Reference} with \textbf{Paraphrased} versions of it. General domain neural paraphrases performed poorly on scientific text. As such, we collect 10 paraphrases from relevant domain experts (each an author of this paper), and incorporate them as few-shot demonstrations for paraphrase generation by GPT-3 \citep{gpt3}. In Appendix \ref{app:gpt3}, we provide more details and show an example. \linebreak

\begin{table}[h]
\centering
\small
\begin{tabular}{l|c|c}
\hline
\textbf{Method} & \textbf{Hyper-Param} & \textbf{Number} \\
\hline
Mask-And-Fill (\emph{Low}) & $m=0.25$ & 10 \\
Mask-And-Fill (\emph{High}) & $m=0.75$ & 10 \\
Swap Intrinsic (\emph{Low}) & $s=0.5$ & 10 \\
Swap Intrinsic (\emph{High}) & $s=1.0$ & 10 \\
Swap Extrinsic (\emph{Low}) & $s=0.5$ & 10 \\
Swap Extrinsic (\emph{High}) & $s=1.0$ & 10 \\ 
Paraphrase & $t=0.7$ & 5 \\
Reference & N/A & 1 \\ \hline
\textbf{Total For Faithfulness} & & \textbf{66} \\ \hline \hline
Diverse Beam (PRIMERA) & $p=1$ & 10 \\
Diverse Beam (LongT5) & $p=1$ & 10 \\ \hline
\textbf{Total For Relevance} & & \textbf{20} \\
\hline
\end{tabular}
\caption{\# of candidates pooled for each training instance. $m$ is \% of noun phrases masked, $s$ \% of entities swapped, and $t$ the softmax temperature for GPT-3. } \label{tab:method-counts}
\vskip -0.1in
\end{table}

\subsection{Candidate Set Details} \label{sec:calibration-analysis}

Table \ref{tab:method-overview} displays the differences between candidate methods at a very basic level, as well as the particular models used for our experiments on long-form scientific summarization. In Table~\ref{tab:method-counts}, we show the number of distinct candidates we produce for each example in the training set by each method / hyper-parameter combination. When calibrating for faithfulness, we select 4 out of 66 possible candidates (2 positive and 2 negative), and for relevance, we select 4 out of 20 possible candidates\footnote{4 is the maximum number which fits in GPU memory on an A100 40GB card, even with a device batch size of one (with gradient accumulation steps) and half precision (fp16).}.




\begin{figure*}[t]
\centering
\includegraphics[width=\linewidth]{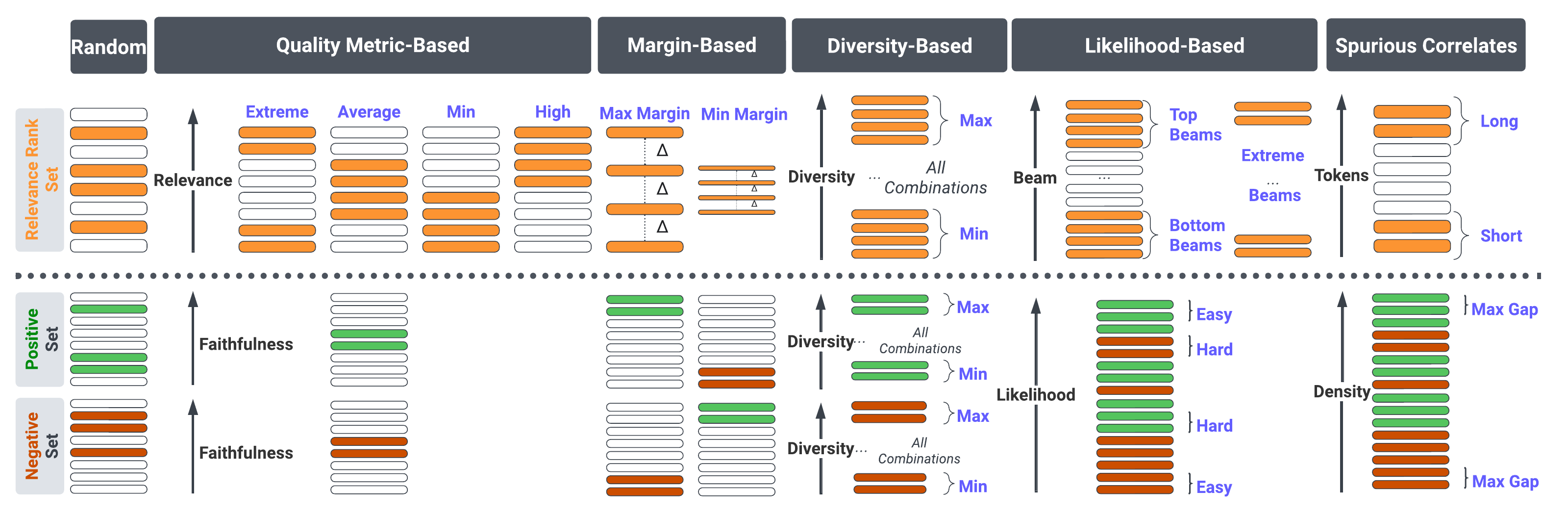}
\caption{Strategies for selecting \textcolor{orange}{rank sets} of size 4 from larger candidate pools for relevance calibration (top half). The bottom half shows similar strategies to form binary contrast sets (2 \textcolor{Green}{positive}, 2 \textcolor{red}{negative}) for faithfulness. Each strategy for the top half of the Figure occupies a row in Table \ref{tab:relevance-results}, while the bottom corresponds to rows in Table \ref{tab:faithfulness-results}. } 
\label{fig:strategies}
\vskip -0.1in
\end{figure*}

\section{Selection Strategies.} \label{sec:strategies}

\paragraph{Problem Statement.} From a large candidate pool, select a target number to be used for \texttt{CT} (2 positives and 2 negatives for faithfulness, and 4 for rank-based relevance). Figure \ref{fig:strategies} graphically reveals the different strategies implemented which are designed to target specific set characteristics. They do not represent optimal or recommended strategies, e.g., a minimum metric gap for faithfulness. In Appendix \ref{app:hypothesis}, we hypothesize as to the specific nature and direction of the impact of the above characteristics on post-calibration summaries.

\paragraph{Random.} For random, for each training instance, we take a random sample without replacement.

\paragraph{Quality-Based.} For quality-based, we rank all candidates by $Rel_{Agg}$ or $Faith_{Agg}$. Then, we select candidates at different extremes of these scales.

\paragraph{Margin-Based.} For relevance ranking, we enumerate all possible subsets of size 4 and compute the average metric margin $Avg(Rel_{Agg}(\hat{S_i}, S) - Rel_{Agg}(\hat{S_{i+1}}, S)), i \in |\bm{\hat{S}}| - 1$. We implement both extremes: one which selects the set with the \texttt{Max Margin}, and its inverse, \texttt{Min Margin}. For faithfulness contrast sets, we either take the most faithful positives and least faithful negatives (\texttt{Max Margin}) or the inverse (\texttt{Min Margin}).

\paragraph{Diversity.} For relevance ranking, we also enumerate all possible subsets of 4 and rank them by their average pairwise inverse self-BLEU score (1 - self-BLEU). We either take the set which has the most \texttt{Max} or \texttt{Min} lexical diversity. We do the same for Faithfulness, except that candidates are selected separately among positive and negative subsets.

\paragraph{Likelihood.} For relevance ranking, we perform selections based on the model's own beam order. We either take the \texttt{Top Beams} (4), \texttt{Bottom Beams} (4), or top 2 and bottom 2 -- \texttt{Extreme Beams}. For faithfulness, we compute the average token-level log likelihood of generating each candidate in the positive and negative sets after \texttt{FT}. Then we either take the \emph{most} likely positives (2) and \emph{least} likely negatives (2) or the \emph{least} likely positives and the \emph{most} likely negatives. For the former, the model is already well-calibrated, which we call \texttt{Easy}. For the latter, confidence and faithfulness are in conflict, which, in comparison, is \texttt{Hard}.

\paragraph{Spurious Correlates.} For relevance, we take the \texttt{Short}est and \texttt{Long}est summaries. For faithfulness, we filter for the \texttt{Max Extractive Gap}--the most \emph{extractive} positives and most \emph{abstractive} negatives (as measured by the extractive density).


\begin{table*}[t!]
\small
\centering
\begin{tabular}{cl|ccc|ccc|ccc}
&  \textbf{Model} & \multicolumn{3}{c}{\textbf{\texttt{Clinical}}} & \multicolumn{3}{c}{\textbf{\texttt{Chemical}}} & \multicolumn{3}{c}{\textbf{\texttt{Biomedical}}} \\ \hline
\multirow{3}{*}{\texttt{\makecell{Relevance \\ Metrics}}} & & \textbf{R1} & \textbf{R2} & \textbf{BS-Ref} & \textbf{R1} & \textbf{R2} & \textbf{BS-Ref} & \textbf{R1} & \textbf{R2} & \textbf{BS-Ref} \\
 & \textbf{PRIMERA} & \textbf{25.15} & \textbf{9.39} & \textbf{83.81} & \textbf{45.47} & \textbf{16.31} & \textbf{86.24} & \textbf{48.01} & \textbf{20.83} & \textbf{86.25} \\
 & \textbf{LongT5} & 24.22 & 8.57 & 83.15 & 42.51 & 14.46 & 85.74 & 44.32 & 17.91 & 85.02 \\ \hline
  \multirow{3}{*}{\texttt{\makecell{Faithful \\ Metrics}}} & & \textbf{Fact.} & \textbf{Bart.} & \textbf{BS-Src} & \textbf{Fact.} & \textbf{Bart.} & \textbf{BS-Src} & \textbf{Fact.} & \textbf{Bart.} & \textbf{BS-Src} \\
 & \textbf{PRIMERA} & 53.29 & -2.92 & \textbf{83.33} & \textbf{85.96} & \textbf{-6.29} & \textbf{88.89} & \textbf{86.91} & \textbf{-3.77} & \textbf{88.54} \\
 & \textbf{LongT5} & \textbf{53.71} & \textbf{-2.88} & 82.84 & 83.25 & -6.36 & 88.70 & 83.62 & -3.89 & 88.31 \\
\end{tabular}
\caption{ Benchmarking PRIMERA and LongT5 models after initial fine-tuning (\texttt{FT}) for relevance and faithfulness. R1, R2, and BS-Ref stand for Rouge-1/2 F1 and BERTScore F1 vis-a-vis reference, respectively. Fact., Bart., and BS-Src stand for FactScore, BARTScore, and BERTScore F1 vis-a-vis the source. Metrics defined in \S \ref{sec:relevance-setup} and \ref{sec:faithful-setup}.}
\label{tab:ft-results}
\end{table*}

\begin{table*}[t!]
\small
\centering
\begin{tabular}{cl|cc|cc|cc||cc}
\multirow{2}{*}{\texttt{\makecell{Selection \\ Type}}} & \multirow{2}{*}{\texttt{\makecell{Selection \\ Strategy}}} & \multicolumn{2}{c}{\texttt{Clinical}} & \multicolumn{2}{c}{\texttt{Chemical}} & \multicolumn{2}{c}{\texttt{Biomedical}} & \multicolumn{2}{c}{\texttt{Dataset Avg.}} \\
& & $REL$ & $FAITH$ & $REL$ & $FAITH$ & $REL$ & $FAITH$ & $REL$ & $FAITH$ \\ \hline
\texttt{Random} & \texttt{-} & .220 & .180 & .081 & \textcolor{red}{-.038} & .028 & .061 & .110 & .068 \\ \hline
\multirow{4}{*}{\texttt{\makecell{Quality \\ Based}}} & \textit{Extreme} & .263 & .152 & .049 & \textcolor{red}{-.168} & .039 & .002 & .117 & \textcolor{red}{-.005} \\
& \textit{Average} & .028 & \textcolor{red}{-.080} & .015 & .056 & .030 & .025 & .024 & .000 \\
& \textit{Min} & .193 & \textcolor{red}{-.022} & .069 & -.049 & .039 & \textcolor{red}{-.012} & .100 & \textcolor{red}{-.027} \\
& \textit{High} & .218 & .095 & .056 & \textcolor{red}{-.029} & .019 & .004 & .098 & .023 \\ \hline
\multirow{2}{*}{\texttt{\makecell{Margin \\ Based}}} & \textit{Max} & .235 & .210 & .062 & .031 & .032 & \textcolor{red}{-.011} & .110 & .077 \\
& \textit{Min} & .158 & \textcolor{red}{-.115} & .028 & .080 & .014 & .015 & .067 & \textcolor{red}{-.007} \\ \hline
\multirow{2}{*}{\texttt{\makecell{Diversity \\ Based}}} & \textit{Max} & .274 & .151 & .054 & \textcolor{red}{-.166} & .015 & \textcolor{red}{-.011} & .114 & \textcolor{red}{-.009} \\
& \textit{Min} & .275 & .091 & \textcolor{red}{-.049} & \textcolor{red}{-.114} & .020 & \textcolor{red}{-.037} & .082 & \textcolor{red}{-.020} \\ \hline
\multirow{3}{*}{\texttt{\makecell{Likelihood \\ Based}}} & \textit{Extreme Beam} & .260 & .140 & .029 & \textcolor{red}{-.158} & .030 & \textcolor{red}{-.008} & .106 & \textcolor{red}{-.009} \\
& \textit{Top Beam} & .287 & .142 & .066 & \textcolor{red}{-.042} & .030 & \textcolor{red}{-.008} & .128 & .031 \\
& \textit{Bottom Beam} & .101 & .125 & .059 & .085 & .025 & \textcolor{red}{-.002} & .062 & .069 \\ \hline
\multirow{2}{*}{\texttt{\makecell{Spurious \\ Correlates}}} & \textit{Max Length} & .255 & .150 & .051 & \textcolor{red}{-.095} & .017 & \textcolor{red}{-.027} & .108 & .009 \\
& \textit{Min Length} & .181 & .243 & .042 & .052 & .033 & .022 & .085 & .106\\ \hline \hline
\multicolumn{2}{c}{\texttt{Avg. Across Strategies}} & .211 & .104 & .044 & \textcolor{red}{-.040} & .027 & .001 & .094 & .022 \\ \hline
\end{tabular}
\caption{ PRIMERA models calibrated to improve relevance. Calibration candidates are pooled from fine-tuned PRIMERA and LongT5 models. $REL$ stands for $Rel_{Agg}$ (from \S \ref{sec:relevance-setup}). $FAITH$ stands for $Faith_{Agg}$ (from \S \ref{sec:faithful-setup}).}
\label{tab:relevance-results}
\end{table*}

\begin{table*}[t!]
\small
\centering
\begin{tabular}{cl|cc|cc|cc||cc}
\multirow{2}{*}{\texttt{\makecell{Selection \\ Type}}} & \multirow{2}{*}{\texttt{\makecell{Selection \\ Strategy}}} & \multicolumn{2}{c}{\texttt{Clinical}} & \multicolumn{2}{c}{\texttt{Chemical}} & \multicolumn{2}{c}{\texttt{Biomedical}} & \multicolumn{2}{c}{\texttt{Dataset Avg.}} \\
& & $REL$ & $FAITH$ & $REL$ & $FAITH$ & $REL$ & $FAITH$ & $REL$ & $FAITH$ \\ \hline
\texttt{Random} & \texttt{-} & \textcolor{red}{-.264} & .133 & \textcolor{red}{-.054} & .085 & .005 & .165 & \textcolor{red}{-.104} & .128 \\ \hline
\texttt{Quality} & \textit{Average} & \textcolor{red}{-.293} & .160 & \textcolor{red}{-.065} & .037 & .010 & .169 & \textcolor{red}{-.116} & .122 \\ \hline
\multirow{2}{*}{\texttt{\makecell{Margin \\ Based}}} & \textit{Max} & \textcolor{red}{-.326} & .313 & \textcolor{red}{-.139} & .011 & \textcolor{red}{-.033} & .018 & \textcolor{red}{-.166} & .114 \\
& \textit{Min} & \textcolor{red}{-.083} & .297 & \textcolor{red}{-.109} & .112 & \textcolor{red}{-.030} & .039 & \textcolor{red}{-.074} & .149 \\ \hline 
\multirow{2}{*}{\texttt{\makecell{Diversity \\ Based}}} & \textit{Max} & .002 & .290 & \textcolor{red}{-.124} & .043 & \textcolor{red}{-.052} & .029 & \textcolor{red}{-.058} & .121 \\
& \textit{Min} & \textcolor{red}{-.039} & .315 & \textcolor{red}{-.040} & .101 & \textcolor{red}{-.043} & .093 & \textcolor{red}{-.041} & .170 \\ \hline
\multirow{2}{*}{\texttt{\makecell{Likelihood \\ Based}}} & \textit{Easy} & .043 & .177 & -.058 & .002 & \textcolor{red}{-.024} & .071 & \textcolor{red}{-.013} & .083 \\
& \textit{Hard} & .071 & .174 & \textcolor{red}{-.233} & .215 & .013 & .147 & \textcolor{red}{-.050} & .179 \\ \hline
\texttt{Spurious} & \textit{Max Extract. Gap} & .044 & .278 & .058 & .046 & \textcolor{red}{-.051} & .067 & .017 & .131 \\ \hline \hline
\multicolumn{2}{c}{\texttt{Avg. Across Strategies}} & \textcolor{red}{-.094} & .237 & \textcolor{red}{-.085} & .072 & \textcolor{red}{-.023} & .089 & \textcolor{red}{-.067} & .133 \\ \hline
\end{tabular}
\caption{PRIMERA models calibrated to improve faithfulness. Contrast sets for calibration are formed from the generation methods in \S \ref{sec:faithful-setup}. $REL$ stands for $Rel_{Agg}$ (from \S \ref{sec:relevance-setup}). $FAITH$ stands for $Faith_{Agg}$ (from \S \ref{sec:faithful-setup}). }
\label{tab:faithfulness-results}
\end{table*}

\section{Results} \label{sec:results}

Please refer to Appendix \ref{app:training-details} for  implementation details on \texttt{FT} and \texttt{CT} training and hyper-parameters.

\subsection{Fine-Tuning} \label{sec:ft-results}


Table \ref{tab:ft-results} shows that PRIMERA outperforms LongT5 across faithfulness and relevance and across datasets\footnote{We note that these our results from own runs. They do not represent results from the PRIMERA and LongT5 papers.}. Relevance and faithfulness are much higher for abstract generation (Chemical and Biomedical) than for clinical summarization, which has highly noisy references. Interestingly, the BARTScore results are lowest for the chemical dataset (-6.29/-6.36 versus -2.92/-2.88 and -3.77/-3.89). This underscores the difference in biomedical versus chemistry-specific papers because the BARTScore model used was trained on the PubMed dataset (\texttt{google/pegasus-pubmed}).


\subsection{Calibration Tuning} \label{sec:calibration-results}

In Tables \ref{tab:relevance-results} and \ref{tab:faithfulness-results}, we report results for relevance, rank-based calibration (\S \ref{sec:relevance-setup}) and faithfulness contrastive learning (\S \ref{sec:faithful-setup}), respectively. $Rel_{Agg}$ and $Faith_{Agg}$ are normalized such that positive values represent standard deviation improvements over fine-tuning, while negative results show a decrease in performance from calibration (marked in \textcolor{red}{red}).

In the following sections, we break down analysis into a \textit{tl;dr}, \textit{evidence}, \textit{explanation}, and potential \textit{implications}, or takeaways, for future research.

Appendix \ref{app:spurious} details the impact of spurious correlates (i.e., length and extractiveness of candidates).

\subsection{The Impact of Reference Quality} \label{sec:dataset-results}

\paragraph{tl;dr.} Relevance and faithfulness calibration offer the most upside when references are noisy.



\paragraph{Evidence.} As detailed in \citet{adams2022learning}, clinical references are often unsupported by the source text. The average across strategies for both Tables \ref{tab:relevance-results} and \ref{tab:faithfulness-results} reveal the largest relative improvement in $Rel_{Agg}$ and $Faith_{Agg}$ for clinical, respectively ($.211$ / $.237$ versus $.044$ / $.072$ and $.027$ / $.089$ for chemical and biomedical abstracts).

\paragraph{Explanation.} For relevance calibration, it is likely that training on model outputs, especially highly extractive ones, dampens some of the noise from variable references. For faithfulness, the rationale is less clear because the reference (and paraphrases of it) form the positive set. Yet, there is an extensive body of work to suggest that training on unfaithful references leads to unfaithful outputs \citep{loss-truncation}, which might make calibrating for faithfulness more impactful.

\paragraph{Implications.} Calibration could be complementary to other methods which address noisy references, such as loss truncation \citep{loss-truncation}, data filtering \citep{narayan-etal-2021-planning, nan-etal-2021-entity}, and reference revision \citep{wan-bansal-2022-factpegasus, adams2022learning}.

\begin{figure}[h]
\centering
\includegraphics[width=\linewidth]{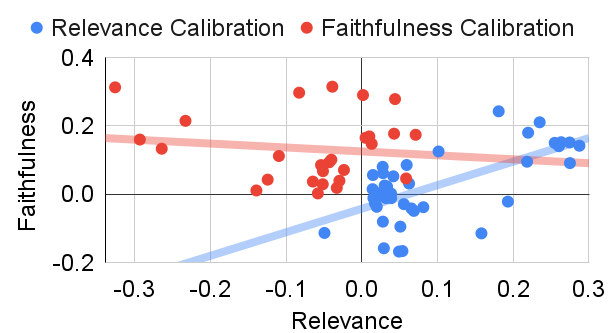}
\caption{A plot of average summary relevance and faithfulness across experiments, which are designed to either improve relevance (blue) or faithfulness (red).} 
\label{fig:rel-versus-faith}
\vskip -0.1in
\end{figure}

\subsection{Relevance and Faithfulness at Odds}

\paragraph{tl;dr.} Relevance and faithfulness share an inverse relationship when calibrating for faithfulness. Research should focus on designing contrast sets that maximize their correlation for joint optimization.

\paragraph{Evidence.} In Figure~\ref{fig:rel-versus-faith}, we plot $Rel_{Agg}$ versus $Faith_{Agg}$ across experiments to measure the tradeoff between relevance and faithfulness. On average, improving faithfulness comes at the cost of relevance, yet the trend is not conclusive. This is validated by previous work which shows a decrease in relevance when models are trained to be more faithful \citep{filippova2020controlled, narayan-etal-2021-planning}. Faithfulness and relevance appear to be positively related when calibrating for relevance. This might be a spurious correlation, however. Model summaries are more extractive than references for each dataset. Including highly extractive summaries as candidates for calibration, in turn, leads to to even more extractive models, as the extractive density of PRIMERA summaries rises from 3.1 / 9.2 / 13.0 after \texttt{FT} to an average of 3.5 / 11.4 / 14.0 for clinical / chemical / biomedical after a round of calibration.

\begin{table}[h]
\centering
\small
\begin{tabular}{l|ccc|c}
\hline
\textbf{System} & \textbf{Int.} & \textbf{Ext.} & \textbf{Total} & \textbf{Rel. Rank} \\
\texttt{FT} & 2.00 & 1.24 & 3.24 & 2.04 \\
\texttt{Most Relevant} & 1.67 & 1.43 & 3.10 & 1.85 \\
\texttt{Most Faithful} & 1.10 & 0.81 & 1.90 & 2.12 \\
\hline
\end{tabular}
\caption{Results from human evaluation on 75 total system summaries from the chemistry test set. Int. and Ext. stand for average intrinsic and extrinsic errors identified. Rel. Rank stands for the average rank assigned by annotators (1-3) with 1 being viewed as the most relevant.} \label{tab:human-eval}
\vskip -0.1in
\end{table}

To see if this relationship is meaningful, we conduct a human evaluation with trained chemists on a random sample of 25 papers from the chemistry test set. For each generated abstract, we ask annotators to separately highlight intrinsic and extrinsic errors, and then to rank each by relevance. We consider abstracts from 3 systems (75 abstracts): the \texttt{Most Relevant} system (according to $Rel_{Agg}$), from relevance calibration (\texttt{Random}), \texttt{Most Faithful } (according to $Faith_{Agg}$) from faithfulness calibration (\texttt{Likelihood - Hard}), and the \texttt{FT} model.

On a small sample, Table \ref{tab:human-eval} confirms what the metrics reveal: an inverse relationship between faithfulness (Int., Ext., Total error counts) and relevance (Rel. Rank). \texttt{Most Faithful} (according to $Faith_{Agg}$) summaries contain the fewest annotated total errors ($1.90$ versus $3.24$ and $3.10$) yet are ranked least relevant (average rank of $2.12$ versus $2.04$ and $1.85$). \texttt{Most Relevant} (according to metrics) achieves the highest relevance ranking from experts ($1.85$ versus $2.04$ / $2.12$) while slightly reducing the number of errors from $FT$: $3.10$ versus $3.10$. On average, there are more intrinsic errors versus extrinsic, which makes sense given how extractive the generated abstracts are. \texttt{Most Relevant} abstracts contain the highest average number of Extrinsic errors ($1.43$ versus $1.24$ and $0.81$), which could stem from the fact that abstracts, as naturally occurring summaries, may introduce external knowledge into the abstracts, for which the \texttt{Most Relevant} may be mimicking.

Please refer to Appendix \ref{app:human} for more details on the annotation protocol and instructions.

\paragraph{Explanation.} From Table \ref{tab:method-metrics}, while references, from a metric perspective, are perfectly relevant, the GPT-3 paraphrases are seen as slightly less relevant (0.9 / 0.94 / 0.92), on average, than the negative methods (0.94 / 0.97 / 0.97) in aggregate). This is likely a by-product of the fact that the negative generation methods selected for this paper involve local corruptions to the reference. The meaning is changed but the word overlap is similar. The GPT-3 paraphrases are prompted with human paraphrases, which involve more substantial re-writing. 



\paragraph{Implications.} Most calibration research is focused on either relevance or faithfulness. We advocate that more papers address them together, since both informativeness and faithfulness are important for real-world systems. Future research could explore joint calibration by intentionally introducing more errors into less relevant summaries.

\begin{table}[h]
\small
\centering
\begin{tabular}{l|ccc|ccc}
 & \multicolumn{3}{c}{\textbf{Average Strategy}} & \multicolumn{3}{c}{\textbf{Max Correlation}} \\
& Rel & Faith & Comb & Rel & Faith & Comb \\ \hline
\textbf{Clin.} & .211 & .104 & .158 & .090 & .325 & .208 \\
\textbf{Chem.} & .044 & -.040 & .007 & .040 & .104 & .158 \\
\textbf{Bio.} & .027 & .001 & .014 & .018 & .025 & .022 \\ \hline
\textbf{Avg.} & .094 & .022 & \textbf{.059} & .049 & .151 & \textbf{.100} \\  \hline
\end{tabular}
\caption{Relevance \texttt{CT} by forming sets which maximize rank correlation between Rel. and Faith. scores improves mean combined (comb.) Rel. and Faith. scores vis-a-vis an average of the strategies shown in Table \ref{tab:relevance-results}.}
\label{tab:both}
\end{table}

As a quick proof of concept, we define a hybrid selection strategy which maximizes the rank correlation between $Agg_{Rel}$ and $Agg_{Faith}$. Table \ref{tab:both} demonstrates that calibrating on these sets leads to positive (pareto) improvements for both metrics. The average improvement in combined metrics across datasets is $.1$, which is greater than an average of the strategies shown in Table \ref{tab:relevance-results} ($.059$).

\begin{figure}[h]
\centering
\includegraphics[width=\linewidth]{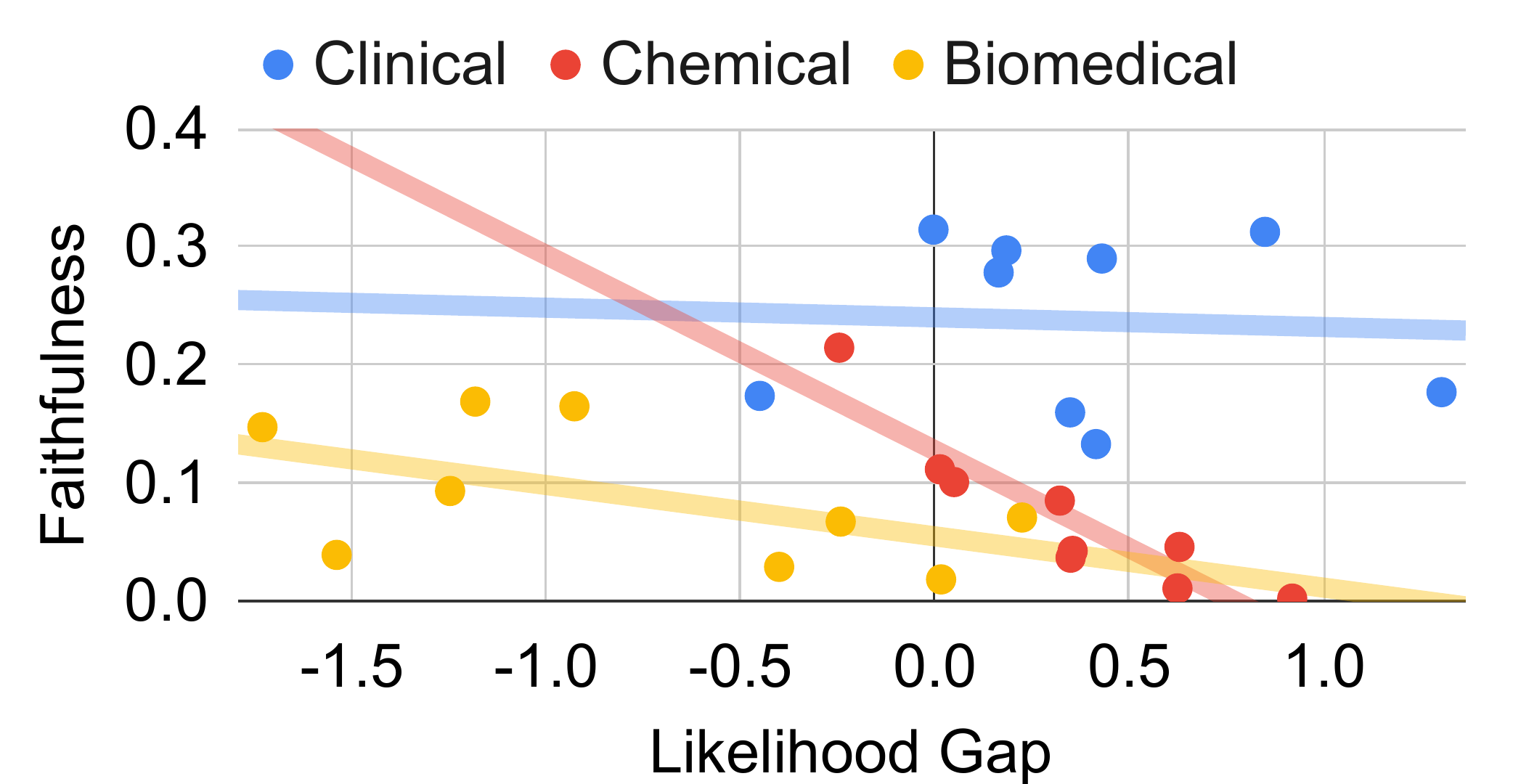}
\caption{A plot comparing the average likelihood gap (difference in log likelihood of generating a positive candidate over a negative pre-calibration) against the average summary faithfulness after calibration.} 
\label{fig:easy-hard}
\vskip -0.1in
\end{figure}

\begin{figure}[h]
\centering
\includegraphics[width=\linewidth]{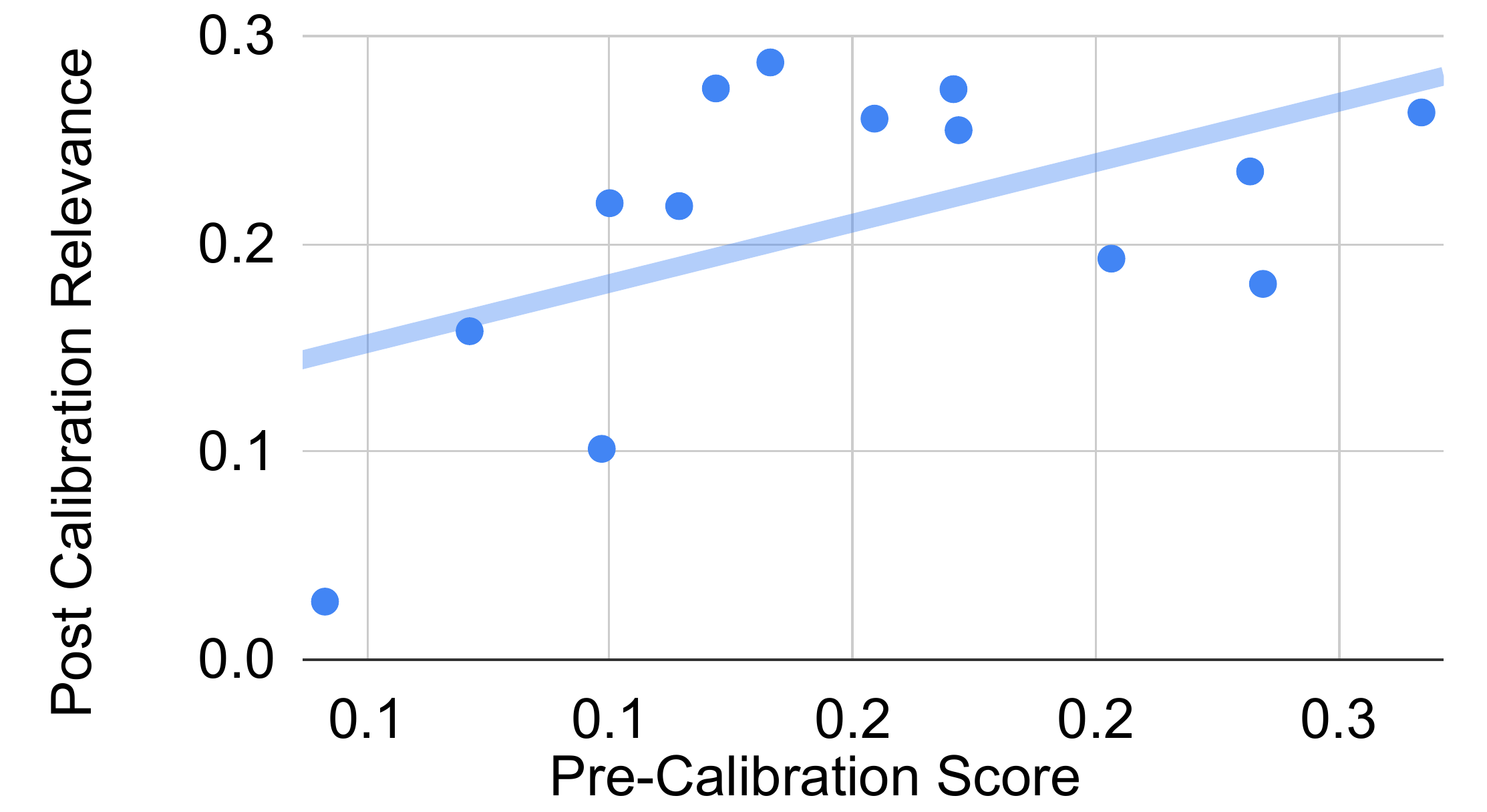}
\caption{A plot which shows average pre-calibration score for each clinical relevance experiment on the x-axis, and the post-calibration relevance on the y-axis.} 
\label{fig:pre-calibration-score}
\vskip -0.1in
\end{figure}

\subsection{On the Dual Role of Surprise} \label{sec:surprise}

\paragraph{tl;dr.} Summaries in sets should be likely under the fine-tuned model. Yet, for relevance, this confidence should mostly already agree with the oracle ranking, while contrastive learning for faithfulness is most effective when the model is surprised.

\paragraph{Evidence.} For relevance, we look at the \texttt{Likelihood} section of Table \ref{tab:relevance-results} and note that, of all strategies, taking the top 4 beams is the most effective (an average of $.128$ across datasets). Taking the bottom beams is one of the worst ($.062$) and taking some from each lies in the middle ($.106$). For faithfulness, we examine  the \texttt{Likelihood} section of Table \ref{tab:faithfulness-results} and note that \texttt{Hard} is the best strategy, on average, across datasets ($.179$ for $Faith_{Agg}$) and \texttt{Easy} is the worst ($-.083$). \texttt{Hard} selects negatives which are most likely under the model, which suggests that contrastive learning for faithfulness is most effective when the model is ``surprised'', i.e., the negative summaries are as likely, if not more, to be generated as the positives.

Across all selection strategies and datasets, we can compute the pre-calibration, average likelihood gap between positives and negatives and regress it against the post-calibration $Faith_{Agg}$ (Figure \ref{fig:easy-hard}). An inverse relationship emerges, especially for chemical dataset (a pearson correlation of $-.91$).

We can run a similar analysis for relevance calibration by computing an average pre-calibration score for each selected set, which we define as the negative spearman correlation coefficient between the model beam and the $Rel_{Agg}$ ranking. It measures the extent to which the model is pre-calibrated from MLE \texttt{FT}. We plot this set statistic against the post-calibration $Agg_{Rel}$ score, as shown in Figure \ref{fig:pre-calibration-score}. The pearson correlation coefficient for the pre-calibration statistic to post-calibration relevance is .52, which is stronger than the correlation of average beam of candidates to relevance (.45).

\begin{figure}[h]
\centering
\includegraphics[width=\linewidth]{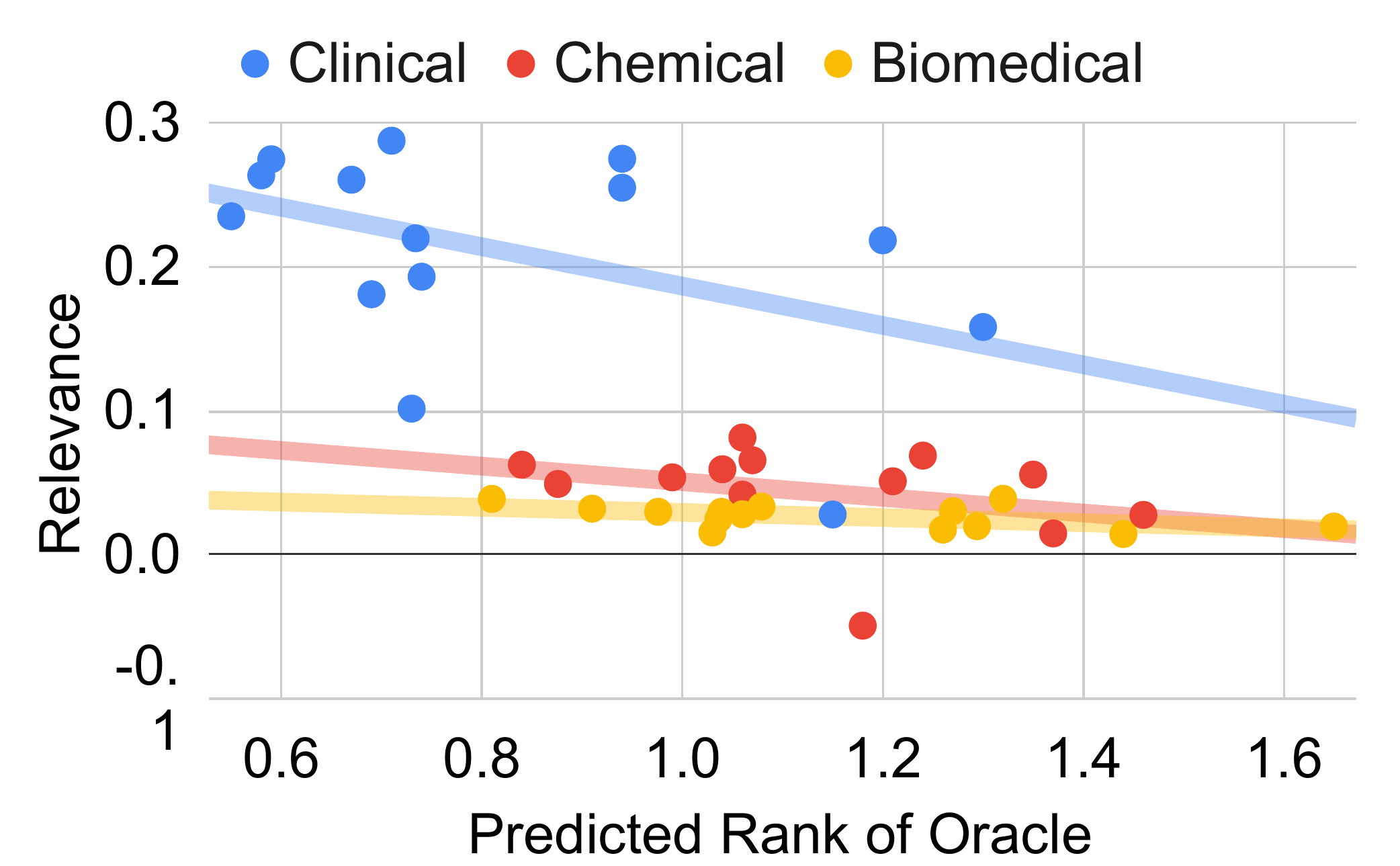}
\caption{A plot showing the impact of calibration performance on downstream performance (relevance). An average rank of 0 reveals a model which always identifies the most relevant summary. The worst score is 3.} 
\label{fig:predicted-rank}
\end{figure}

We can also link the model's ranking ability \emph{after} calibration to the post-calibration relevance. In other words, does it matter how well the model can rank candidates given that, when used for inference, it generates a single candidate? Figure \ref{fig:predicted-rank} shows that a well calibrated model is a better generator due to an inverse relationship between the predicted rank of the top ranked candidate (x-axis) and the average post-calibration $Rel_{Agg}$ score (y-axis).

Taken together, the results suggest that an optimal rank set for relevance is one that is fairly calibrated before \texttt{CT} and well-calibrated after \texttt{CT}.

\paragraph{Explanation.} A possible explanation for this conflicting evidence is a difference in objectives. As in \citet{liu-etal-2022-brio}, the relevance ordering is directly calibrated to log likelihood of outputs, whereas for faithfulness, we contrast binary positives and negatives in latent space. For the former, large parameter updates from the ranking loss directly affect the generation behavior of the model, which \emph{may} push outputs further away from the MLE optimum.

\paragraph{Implications.} The results suggest it might be preferable to \emph{surprise} for faithfulness calibration yet \emph{confirm} for relevance calibration. Yet, further work is necessary to assess whether this behavior is attributable to the objective or the metric. 


\begin{figure}[h]
\centering
\includegraphics[width=\linewidth]{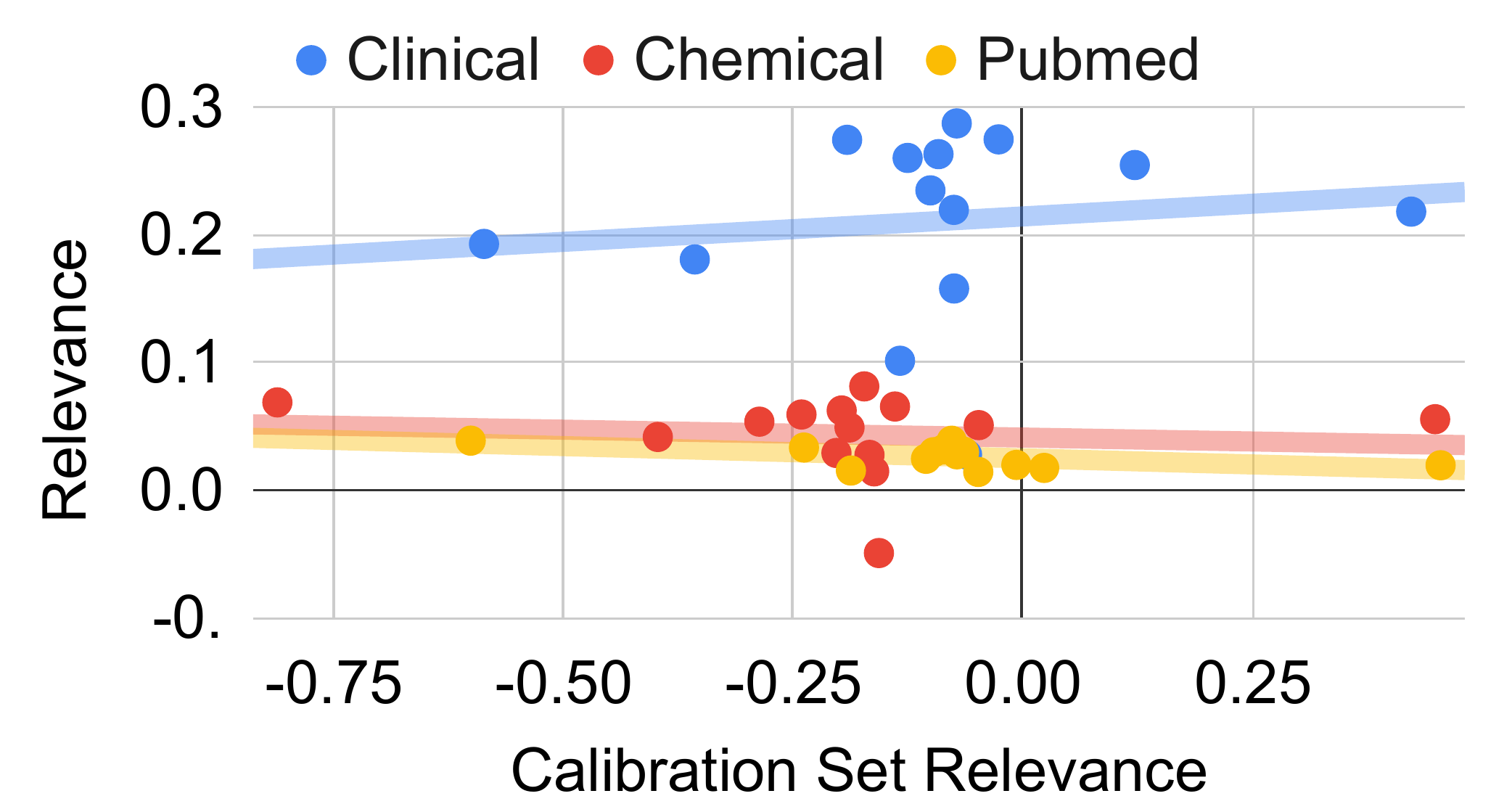}
\caption{The impact of the average relevance of calibration candidates on downstream summary relevance.} 
\label{fig:absolute}
\end{figure}

\begin{figure}[h]
\centering
\includegraphics[width=\linewidth]{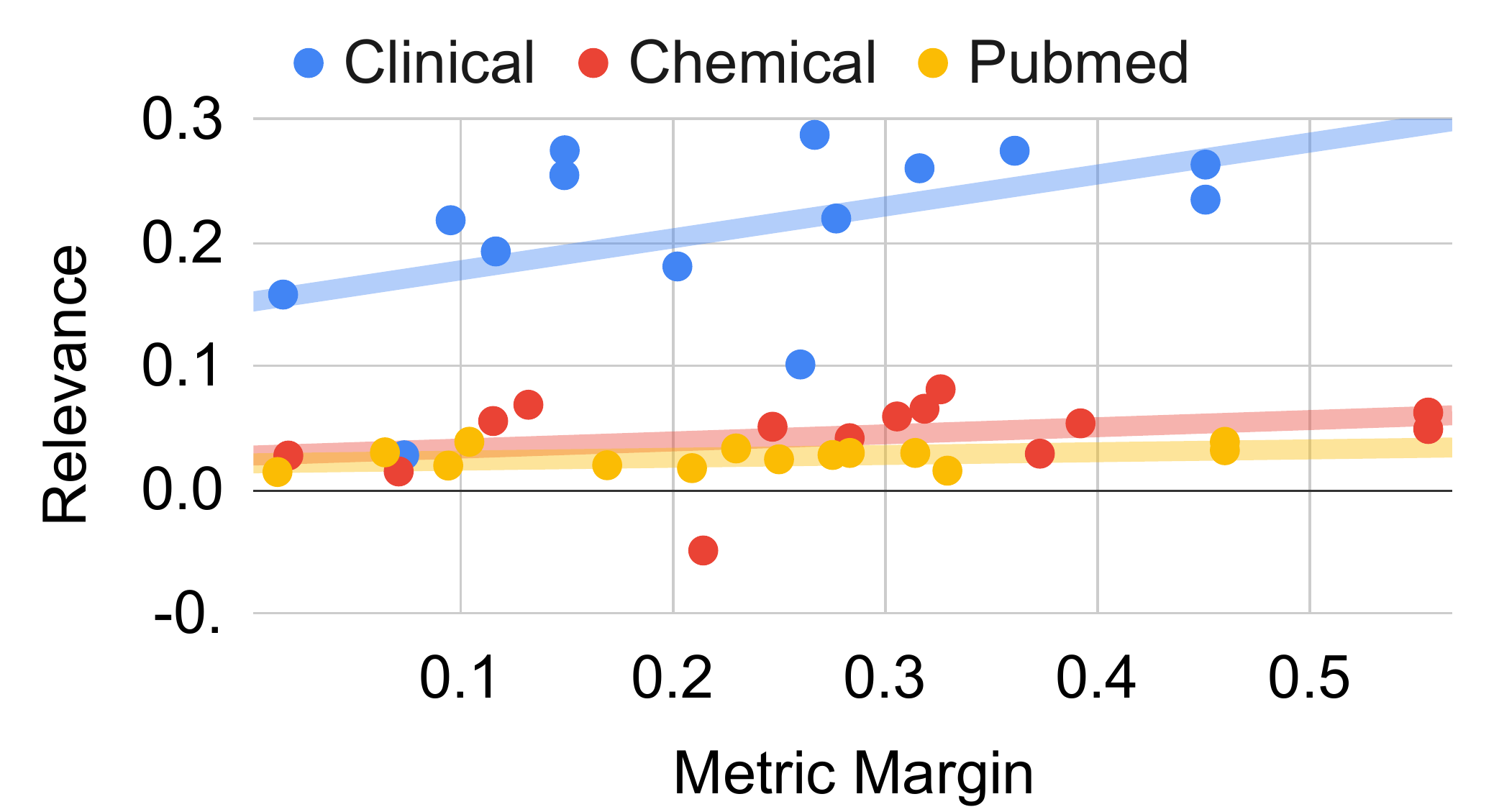}
\caption{The impact of the average metric-wise margin ($Rel_{Agg}$) between calibration \emph{candidates} on the relevance of downstream model outputs after calibration.}
\label{fig:margin}
\end{figure}

\subsection{Margin over Absolute}

\paragraph{tl;dr.} For relevance training, the presence of a large metric margin between candidate summaries appears to be more impactful to downstream performance than the overall relevance of the set.

\paragraph{Evidence.} Based on Table~\ref{tab:relevance-results} for \texttt{Quality Based} Avg. Across Strategies, no clear-cut trend exists between $Rel_{Agg}$ and absolute relevance values: $.117 / .024 / .100 / .098 $ for \texttt{Extreme}, \texttt{Average}, \texttt{Min}, and \texttt{High}, respectively. For \texttt{Margin Based}, which targets the relative values, \texttt{Max} outperforms $.110$ over $.067$. To better uncover any trends, we separately plot the average set relevance (absolute value), and the Margin Gap (relative values), against downstream $Rel_{Agg}$ for each run (row in Table~\ref{tab:relevance-results}) in Figures~\ref{fig:absolute} and \ref{fig:margin}. Figure~\ref{fig:margin} shows a positive correlation between margin gap and downstream $Rel_{Agg}$ across datasets (pearson correlation of $.48$, $.29$, and $.38$ for clinical, chemical, and biomedical, respectively). The relationship in Figure \ref{fig:absolute} is less consistent, as it is positive for clinical ($.12$ correlation), yet negative for chemical ($-.10$) and biomedical ($-.51$). We connect margins to diversity in Appendix \ref{sec:connections}.

\paragraph{Implications.} Diversity may help calibration with increased exploration and smooth out some noise from ROUGE / BERTScore defined rankings. Although \citet{zhao2022calibrating} find consistently better performance using regular beam search over diverse beam search, the opposite may hold true for longer tasks with larger output search spaces.
\section{Conclusion}

In this paper, we explore what makes an effective calibration set for both relevance and faithfulness tuning. To do so, we create large candidate pools for calibration and design strategies which systematically target set characterstics. We then analyze trends between these characteristics and downstream performance. Our analysis is intended to serve as a guide for subsequent research when designing methods to form synthetic candidates, as well as motivation to jointly consider relevance and faithfulness for calibration, given their covariance and the importance of both to real-world systems.

\section{Limitations}

As we cannot control for all confounding variables when examining the correlates of the most effective contrast sets, we only claim to identify trends, not causality, between calibration set characteristics and downstream performance. For instance, the top beams, on average, have higher relevance. As such, for each strategy, we record \emph{all} key set characteristics and focus our analysis on observing trends between set characteristic values and downstream performance across \emph{all} experiments, not simply within each \texttt{Selection Type}.


\bibliography{anthology,custom}

\begin{thebibliography}{78}
\expandafter\ifx\csname natexlab\endcsname\relax\def\natexlab#1{#1}\fi

\bibitem[{Adams et~al.(2021)Adams, Alsentzer, Ketenci, Zucker, and
  Elhadad}]{adams-etal-2021-whats}
Griffin Adams, Emily Alsentzer, Mert Ketenci, Jason Zucker, and No{\'e}mie
  Elhadad. 2021.
\newblock \href {https://doi.org/10.18653/v1/2021.naacl-main.382} {What{'}s in
  a summary? laying the groundwork for advances in hospital-course
  summarization}.
\newblock In \emph{Proceedings of the 2021 Conference of the North American
  Chapter of the Association for Computational Linguistics: Human Language
  Technologies}, pages 4794--4811, Online. Association for Computational
  Linguistics.

\bibitem[{Adams et~al.(2022)Adams, Shing, Sun, Winestock, McKeown, and
  Elhadad}]{adams2022learning}
Griffin Adams, Han-Chin Shing, Qing Sun, Christopher Winestock, Kathleen
  McKeown, and No{\'e}mie Elhadad. 2022.
\newblock Learning to revise references for faithful summarization.
\newblock \emph{ArXiv}, abs/2204.10290.

\bibitem[{Ainslie et~al.(2020)Ainslie, Ontanon, Alberti, Cvicek, Fisher, Pham,
  Ravula, Sanghai, Wang, and Yang}]{ainslie-etal-2020-etc}
Joshua Ainslie, Santiago Ontanon, Chris Alberti, Vaclav Cvicek, Zachary Fisher,
  Philip Pham, Anirudh Ravula, Sumit Sanghai, Qifan Wang, and Li~Yang. 2020.
\newblock \href {https://doi.org/10.18653/v1/2020.emnlp-main.19} {{ETC}:
  Encoding long and structured inputs in transformers}.
\newblock In \emph{Proceedings of the 2020 Conference on Empirical Methods in
  Natural Language Processing (EMNLP)}, pages 268--284, Online. Association for
  Computational Linguistics.

\bibitem[{Alihosseini et~al.(2019)Alihosseini, Montahaei, and
  Soleymani~Baghshah}]{alihosseini-etal-2019-jointly}
Danial Alihosseini, Ehsan Montahaei, and Mahdieh Soleymani~Baghshah. 2019.
\newblock \href {https://doi.org/10.18653/v1/W19-2311} {Jointly measuring
  diversity and quality in text generation models}.
\newblock In \emph{Proceedings of the Workshop on Methods for Optimizing and
  Evaluating Neural Language Generation}, pages 90--98, Minneapolis, Minnesota.
  Association for Computational Linguistics.

\bibitem[{Beltagy et~al.(2020)Beltagy, Peters, and
  Cohan}]{beltagy2020longformer}
Iz~Beltagy, Matthew~E Peters, and Arman Cohan. 2020.
\newblock Longformer: The long-document transformer.
\newblock \emph{arXiv preprint arXiv:2004.05150}.

\bibitem[{Bodenreider(2004)}]{bodenreider2004unified}
Olivier Bodenreider. 2004.
\newblock The unified medical language system (umls): integrating biomedical
  terminology.
\newblock \emph{Nucleic acids research}, 32(suppl\_1):D267--D270.

\bibitem[{Brown et~al.(2020)Brown, Mann, Ryder, Subbiah, Kaplan, Dhariwal,
  Neelakantan, Shyam, Sastry, Askell et~al.}]{gpt3}
Tom Brown, Benjamin Mann, Nick Ryder, Melanie Subbiah, Jared~D Kaplan, Prafulla
  Dhariwal, Arvind Neelakantan, Pranav Shyam, Girish Sastry, Amanda Askell,
  et~al. 2020.
\newblock Language models are few-shot learners.
\newblock \emph{Advances in neural information processing systems},
  33:1877--1901.

\bibitem[{Cao et~al.(2020)Cao, Dong, Wu, and Cheung}]{cao-etal-2020-factual}
Meng Cao, Yue Dong, Jiapeng Wu, and Jackie Chi~Kit Cheung. 2020.
\newblock \href {https://doi.org/10.18653/v1/2020.emnlp-main.506} {Factual
  error correction for abstractive summarization models}.
\newblock In \emph{Proceedings of the 2020 Conference on Empirical Methods in
  Natural Language Processing (EMNLP)}, pages 6251--6258, Online. Association
  for Computational Linguistics.

\bibitem[{Cao and Wang(2021{\natexlab{a}})}]{cliff}
Shuyang Cao and Lu~Wang. 2021{\natexlab{a}}.
\newblock \href {https://doi.org/10.18653/v1/2021.emnlp-main.532} {{CLIFF}:
  Contrastive learning for improving faithfulness and factuality in abstractive
  summarization}.
\newblock In \emph{Proceedings of the 2021 Conference on Empirical Methods in
  Natural Language Processing}, pages 6633--6649, Online and Punta Cana,
  Dominican Republic. Association for Computational Linguistics.

\bibitem[{Cao and Wang(2021{\natexlab{b}})}]{cao-wang-2021-cliff}
Shuyang Cao and Lu~Wang. 2021{\natexlab{b}}.
\newblock \href {https://doi.org/10.18653/v1/2021.emnlp-main.532} {{CLIFF}:
  Contrastive learning for improving faithfulness and factuality in abstractive
  summarization}.
\newblock In \emph{Proceedings of the 2021 Conference on Empirical Methods in
  Natural Language Processing}, pages 6633--6649, Online and Punta Cana,
  Dominican Republic. Association for Computational Linguistics.

\bibitem[{Cao et~al.(2018)Cao, Wei, Li, and Li}]{cao2018faithful}
Ziqiang Cao, Furu Wei, Wenjie Li, and Sujian Li. 2018.
\newblock Faithful to the original: Fact aware neural abstractive
  summarization.
\newblock In \emph{Proceedings of the AAAI Conference on Artificial
  Intelligence}, volume~32.

\bibitem[{Chen et~al.(2021)Chen, Zhang, Sone, and
  Roth}]{chen-etal-2021-improving}
Sihao Chen, Fan Zhang, Kazoo Sone, and Dan Roth. 2021.
\newblock \href {https://doi.org/10.18653/v1/2021.naacl-main.475} {Improving
  faithfulness in abstractive summarization with contrast candidate generation
  and selection}.
\newblock In \emph{Proceedings of the 2021 Conference of the North American
  Chapter of the Association for Computational Linguistics: Human Language
  Technologies}, pages 5935--5941, Online. Association for Computational
  Linguistics.

\bibitem[{Chintagunta et~al.(2021)Chintagunta, Katariya, Amatriain, and
  Kannan}]{chintagunta-etal-2021-medically}
Bharath Chintagunta, Namit Katariya, Xavier Amatriain, and Anitha Kannan. 2021.
\newblock \href {https://doi.org/10.18653/v1/2021.nlpmc-1.9} {Medically aware
  {GPT}-3 as a data generator for medical dialogue summarization}.
\newblock In \emph{Proceedings of the Second Workshop on Natural Language
  Processing for Medical Conversations}, pages 66--76, Online. Association for
  Computational Linguistics.

\bibitem[{Cohan et~al.(2018)Cohan, Dernoncourt, Kim, Bui, Kim, Chang, and
  Goharian}]{cohan-etal-2018-discourse}
Arman Cohan, Franck Dernoncourt, Doo~Soon Kim, Trung Bui, Seokhwan Kim, Walter
  Chang, and Nazli Goharian. 2018.
\newblock \href {https://doi.org/10.18653/v1/N18-2097} {A discourse-aware
  attention model for abstractive summarization of long documents}.
\newblock In \emph{Proceedings of the 2018 Conference of the North {A}merican
  Chapter of the Association for Computational Linguistics: Human Language
  Technologies, Volume 2 (Short Papers)}, pages 615--621, New Orleans,
  Louisiana. Association for Computational Linguistics.

\bibitem[{Deng et~al.(2021{\natexlab{a}})Deng, Tan, Liu, Xing, and
  Hu}]{deng-etal-2021-compression}
Mingkai Deng, Bowen Tan, Zhengzhong Liu, Eric Xing, and Zhiting Hu.
  2021{\natexlab{a}}.
\newblock \href {https://doi.org/10.18653/v1/2021.emnlp-main.599} {Compression,
  transduction, and creation: A unified framework for evaluating natural
  language generation}.
\newblock In \emph{Proceedings of the 2021 Conference on Empirical Methods in
  Natural Language Processing}, pages 7580--7605, Online and Punta Cana,
  Dominican Republic. Association for Computational Linguistics.

\bibitem[{Deng et~al.(2021{\natexlab{b}})Deng, Tan, Liu, Xing, and Hu}]{ctc}
Mingkai Deng, Bowen Tan, Zhengzhong Liu, Eric Xing, and Zhiting Hu.
  2021{\natexlab{b}}.
\newblock \href {https://doi.org/10.18653/v1/2021.emnlp-main.599} {Compression,
  transduction, and creation: A unified framework for evaluating natural
  language generation}.
\newblock In \emph{Proceedings of the 2021 Conference on Empirical Methods in
  Natural Language Processing}, pages 7580--7605, Online and Punta Cana,
  Dominican Republic. Association for Computational Linguistics.

\bibitem[{DeYoung et~al.(2021)DeYoung, Beltagy, van Zuylen, Kuehl, and
  Wang}]{deyoung-etal-2021-ms}
Jay DeYoung, Iz~Beltagy, Madeleine van Zuylen, Bailey Kuehl, and Lucy Wang.
  2021.
\newblock \href {https://doi.org/10.18653/v1/2021.emnlp-main.594} {{MS}{\^{}}2:
  Multi-document summarization of medical studies}.
\newblock In \emph{Proceedings of the 2021 Conference on Empirical Methods in
  Natural Language Processing}, pages 7494--7513, Online and Punta Cana,
  Dominican Republic. Association for Computational Linguistics.

\bibitem[{Durmus et~al.(2022)Durmus, Ladhak, and
  Hashimoto}]{durmus-etal-2022-spurious}
Esin Durmus, Faisal Ladhak, and Tatsunori Hashimoto. 2022.
\newblock \href {https://doi.org/10.18653/v1/2022.acl-long.102} {Spurious
  correlations in reference-free evaluation of text generation}.
\newblock In \emph{Proceedings of the 60th Annual Meeting of the Association
  for Computational Linguistics (Volume 1: Long Papers)}, pages 1443--1454,
  Dublin, Ireland. Association for Computational Linguistics.

\bibitem[{Fabbri et~al.(2021{\natexlab{a}})Fabbri, Han, Li, Li, Ghazvininejad,
  Joty, Radev, and Mehdad}]{fabbri-etal-2021-improving}
Alexander Fabbri, Simeng Han, Haoyuan Li, Haoran Li, Marjan Ghazvininejad,
  Shafiq Joty, Dragomir Radev, and Yashar Mehdad. 2021{\natexlab{a}}.
\newblock \href {https://doi.org/10.18653/v1/2021.naacl-main.57} {Improving
  zero and few-shot abstractive summarization with intermediate fine-tuning and
  data augmentation}.
\newblock In \emph{Proceedings of the 2021 Conference of the North American
  Chapter of the Association for Computational Linguistics: Human Language
  Technologies}, pages 704--717, Online. Association for Computational
  Linguistics.

\bibitem[{Fabbri et~al.(2021{\natexlab{b}})Fabbri, Kry{\'s}ci{\'n}ski, McCann,
  Xiong, Socher, and Radev}]{fabbri-etal-2021-summeval}
Alexander~R. Fabbri, Wojciech Kry{\'s}ci{\'n}ski, Bryan McCann, Caiming Xiong,
  Richard Socher, and Dragomir Radev. 2021{\natexlab{b}}.
\newblock \href {https://doi.org/10.1162/tacl_a_00373} {{S}umm{E}val:
  Re-evaluating summarization evaluation}.
\newblock \emph{Transactions of the Association for Computational Linguistics},
  9:391--409.

\bibitem[{Filippova(2020)}]{filippova2020controlled}
Katja Filippova. 2020.
\newblock Controlled hallucinations: Learning to generate faithfully from noisy
  data.
\newblock \emph{arXiv preprint arXiv:2010.05873}.

\bibitem[{Gehrmann et~al.(2022)Gehrmann, Clark, and
  Sellam}]{gehrmann2022repairing}
Sebastian Gehrmann, Elizabeth Clark, and Thibault Sellam. 2022.
\newblock Repairing the cracked foundation: A survey of obstacles in evaluation
  practices for generated text.
\newblock \emph{arXiv preprint arXiv:2202.06935}.

\bibitem[{Goyal and Durrett(2020)}]{goyal-durrett-2020-neural}
Tanya Goyal and Greg Durrett. 2020.
\newblock \href {https://doi.org/10.18653/v1/2020.acl-main.22} {Neural
  syntactic preordering for controlled paraphrase generation}.
\newblock In \emph{Proceedings of the 58th Annual Meeting of the Association
  for Computational Linguistics}, pages 238--252, Online. Association for
  Computational Linguistics.

\bibitem[{Goyal and Durrett(2021)}]{goyal-durrett-2021-annotating}
Tanya Goyal and Greg Durrett. 2021.
\newblock \href {https://doi.org/10.18653/v1/2021.naacl-main.114} {Annotating
  and modeling fine-grained factuality in summarization}.
\newblock In \emph{Proceedings of the 2021 Conference of the North American
  Chapter of the Association for Computational Linguistics: Human Language
  Technologies}, pages 1449--1462, Online. Association for Computational
  Linguistics.

\bibitem[{Grusky et~al.(2018)Grusky, Naaman, and Artzi}]{grusky2018newsroom}
Max Grusky, Mor Naaman, and Yoav Artzi. 2018.
\newblock \href {https://doi.org/10.18653/v1/N18-1065} {{N}ewsroom: A dataset
  of 1.3 million summaries with diverse extractive strategies}.
\newblock In \emph{Proceedings of the 2018 Conference of the North {A}merican
  Chapter of the Association for Computational Linguistics: Human Language
  Technologies, Volume 1 (Long Papers)}, pages 708--719, New Orleans,
  Louisiana. Association for Computational Linguistics.

\bibitem[{Gunel et~al.(2020)Gunel, Du, Conneau, and
  Stoyanov}]{gunel2020supervised}
Beliz Gunel, Jingfei Du, Alexis Conneau, and Ves Stoyanov. 2020.
\newblock Supervised contrastive learning for pre-trained language model
  fine-tuning.
\newblock \emph{arXiv preprint arXiv:2011.01403}.

\bibitem[{Guo et~al.(2022)Guo, Ainslie, Uthus, Ontanon, Ni, Sung, and
  Yang}]{longt5}
Mandy Guo, Joshua Ainslie, David Uthus, Santiago Ontanon, Jianmo Ni, Yun-Hsuan
  Sung, and Yinfei Yang. 2022.
\newblock \href {https://doi.org/10.18653/v1/2022.findings-naacl.55}
  {{L}ong{T}5: {E}fficient text-to-text transformer for long sequences}.
\newblock In \emph{Findings of the Association for Computational Linguistics:
  NAACL 2022}, pages 724--736, Seattle, United States. Association for
  Computational Linguistics.

\bibitem[{Gupta et~al.(2021)Gupta, Bharti, Nokhiz, and
  Karnick}]{gupta-etal-2021-sumpubmed}
Vivek Gupta, Prerna Bharti, Pegah Nokhiz, and Harish Karnick. 2021.
\newblock \href {https://doi.org/10.18653/v1/2021.acl-srw.30} {{SumPubMed}:
  Summarization dataset of {P}ub{M}ed scientific articles}.
\newblock In \emph{Proceedings of the 59th Annual Meeting of the Association
  for Computational Linguistics and the 11th International Joint Conference on
  Natural Language Processing: Student Research Workshop}, pages 292--303,
  Online. Association for Computational Linguistics.

\bibitem[{Holtzman et~al.(2019)Holtzman, Buys, Du, Forbes, and
  Choi}]{holtzman2019curious}
Ari Holtzman, Jan Buys, Li~Du, Maxwell Forbes, and Yejin Choi. 2019.
\newblock The curious case of neural text degeneration.
\newblock \emph{arXiv preprint arXiv:1904.09751}.

\bibitem[{Johnson et~al.(2016)Johnson, Pollard, Shen, Li-Wei, Feng, Ghassemi,
  Moody, Szolovits, Celi, and Mark}]{johnson2016mimic}
Alistair~EW Johnson, Tom~J Pollard, Lu~Shen, H~Lehman Li-Wei, Mengling Feng,
  Mohammad Ghassemi, Benjamin Moody, Peter Szolovits, Leo~Anthony Celi, and
  Roger~G Mark. 2016.
\newblock Mimic-iii, a freely accessible critical care database.
\newblock \emph{Scientific data}, 3(1):1--9.

\bibitem[{Kang and Hashimoto(2020)}]{loss-truncation}
Daniel Kang and Tatsunori~B. Hashimoto. 2020.
\newblock \href {https://doi.org/10.18653/v1/2020.acl-main.66} {Improved
  natural language generation via loss truncation}.
\newblock In \emph{Proceedings of the 58th Annual Meeting of the Association
  for Computational Linguistics}, pages 718--731, Online. Association for
  Computational Linguistics.

\bibitem[{Khosla et~al.(2020)Khosla, Teterwak, Wang, Sarna, Tian, Isola,
  Maschinot, Liu, and Krishnan}]{khosla2020supervised}
Prannay Khosla, Piotr Teterwak, Chen Wang, Aaron Sarna, Yonglong Tian, Phillip
  Isola, Aaron Maschinot, Ce~Liu, and Dilip Krishnan. 2020.
\newblock Supervised contrastive learning.
\newblock \emph{Advances in Neural Information Processing Systems},
  33:18661--18673.

\bibitem[{Kim et~al.(2019)Kim, Lee, So, Jeon, Jeong, Choi, Yoon, Sung, and
  Kang}]{kim2019neural}
Donghyeon Kim, Jinhyuk Lee, Chan~Ho So, Hwisang Jeon, Minbyul Jeong, Yonghwa
  Choi, Wonjin Yoon, Mujeen Sung, and Jaewoo Kang. 2019.
\newblock A neural named entity recognition and multi-type normalization tool
  for biomedical text mining.
\newblock \emph{IEEE Access}, 7:73729--73740.

\bibitem[{Krishna et~al.(2021)Krishna, Khosla, Bigham, and
  Lipton}]{krishna-etal-2021-generating}
Kundan Krishna, Sopan Khosla, Jeffrey Bigham, and Zachary~C. Lipton. 2021.
\newblock \href {https://doi.org/10.18653/v1/2021.acl-long.384} {Generating
  {SOAP} notes from doctor-patient conversations using modular summarization
  techniques}.
\newblock In \emph{Proceedings of the 59th Annual Meeting of the Association
  for Computational Linguistics and the 11th International Joint Conference on
  Natural Language Processing (Volume 1: Long Papers)}, pages 4958--4972,
  Online. Association for Computational Linguistics.

\bibitem[{Kryscinski et~al.(2020)Kryscinski, McCann, Xiong, and
  Socher}]{factcc}
Wojciech Kryscinski, Bryan McCann, Caiming Xiong, and Richard Socher. 2020.
\newblock \href {https://doi.org/10.18653/v1/2020.emnlp-main.750} {Evaluating
  the factual consistency of abstractive text summarization}.
\newblock In \emph{Proceedings of the 2020 Conference on Empirical Methods in
  Natural Language Processing (EMNLP)}, pages 9332--9346, Online. Association
  for Computational Linguistics.

\bibitem[{Ladhak et~al.(2022)Ladhak, Durmus, He, Cardie, and
  McKeown}]{ladhak-etal-2022-faithful}
Faisal Ladhak, Esin Durmus, He~He, Claire Cardie, and Kathleen McKeown. 2022.
\newblock \href {https://doi.org/10.18653/v1/2022.acl-long.100} {Faithful or
  extractive? on mitigating the faithfulness-abstractiveness trade-off in
  abstractive summarization}.
\newblock In \emph{Proceedings of the 60th Annual Meeting of the Association
  for Computational Linguistics (Volume 1: Long Papers)}, pages 1410--1421,
  Dublin, Ireland. Association for Computational Linguistics.

\bibitem[{Lebanoff et~al.(2019)Lebanoff, Muchovej, Dernoncourt, Kim, Kim,
  Chang, and Liu}]{lebanoff-etal-2019-analyzing}
Logan Lebanoff, John Muchovej, Franck Dernoncourt, Doo~Soon Kim, Seokhwan Kim,
  Walter Chang, and Fei Liu. 2019.
\newblock \href {https://doi.org/10.18653/v1/D19-5413} {Analyzing sentence
  fusion in abstractive summarization}.
\newblock In \emph{Proceedings of the 2nd Workshop on New Frontiers in
  Summarization}, pages 104--110, Hong Kong, China. Association for
  Computational Linguistics.

\bibitem[{Lee et~al.(2022)Lee, Yoo, Park, Lee, and Jung}]{lee-etal-2022-masked}
Hwanhee Lee, Kang~Min Yoo, Joonsuk Park, Hwaran Lee, and Kyomin Jung. 2022.
\newblock \href {https://doi.org/10.18653/v1/2022.findings-naacl.76} {Masked
  summarization to generate factually inconsistent summaries for improved
  factual consistency checking}.
\newblock In \emph{Findings of the Association for Computational Linguistics:
  NAACL 2022}, pages 1019--1030, Seattle, United States. Association for
  Computational Linguistics.

\bibitem[{Lin(2004)}]{lin2004rouge}
Chin-Yew Lin. 2004.
\newblock Rouge: A package for automatic evaluation of summaries.
\newblock In \emph{Text summarization branches out}, pages 74--81.

\bibitem[{Liu and Liu(2021{\natexlab{a}})}]{simcls}
Yixin Liu and Pengfei Liu. 2021{\natexlab{a}}.
\newblock \href {https://doi.org/10.18653/v1/2021.acl-short.135} {{S}im{CLS}: A
  simple framework for contrastive learning of abstractive summarization}.
\newblock In \emph{Proceedings of the 59th Annual Meeting of the Association
  for Computational Linguistics and the 11th International Joint Conference on
  Natural Language Processing (Volume 2: Short Papers)}, pages 1065--1072,
  Online. Association for Computational Linguistics.

\bibitem[{Liu and Liu(2021{\natexlab{b}})}]{liu-liu-2021-simcls}
Yixin Liu and Pengfei Liu. 2021{\natexlab{b}}.
\newblock \href {https://doi.org/10.18653/v1/2021.acl-short.135} {{S}im{CLS}: A
  simple framework for contrastive learning of abstractive summarization}.
\newblock In \emph{Proceedings of the 59th Annual Meeting of the Association
  for Computational Linguistics and the 11th International Joint Conference on
  Natural Language Processing (Volume 2: Short Papers)}, pages 1065--1072,
  Online. Association for Computational Linguistics.

\bibitem[{Liu et~al.(2022)Liu, Liu, Radev, and Neubig}]{liu-etal-2022-brio}
Yixin Liu, Pengfei Liu, Dragomir Radev, and Graham Neubig. 2022.
\newblock \href {https://doi.org/10.18653/v1/2022.acl-long.207} {{BRIO}:
  Bringing order to abstractive summarization}.
\newblock In \emph{Proceedings of the 60th Annual Meeting of the Association
  for Computational Linguistics (Volume 1: Long Papers)}, pages 2890--2903,
  Dublin, Ireland. Association for Computational Linguistics.

\bibitem[{Lopez(2009)}]{lopez2009grobid}
Patrice Lopez. 2009.
\newblock Grobid: Combining automatic bibliographic data recognition and term
  extraction for scholarship publications.
\newblock In \emph{International conference on theory and practice of digital
  libraries}, pages 473--474. Springer.

\bibitem[{Lu et~al.(2020)Lu, Dong, and Charlin}]{lu-etal-2020-multi-xscience}
Yao Lu, Yue Dong, and Laurent Charlin. 2020.
\newblock \href {https://doi.org/10.18653/v1/2020.emnlp-main.648}
  {Multi-{XS}cience: A large-scale dataset for extreme multi-document
  summarization of scientific articles}.
\newblock In \emph{Proceedings of the 2020 Conference on Empirical Methods in
  Natural Language Processing (EMNLP)}, pages 8068--8074, Online. Association
  for Computational Linguistics.

\bibitem[{Marcinkiewicz(1994)}]{marcinkiewicz1994building}
Mary~Ann Marcinkiewicz. 1994.
\newblock Building a large annotated corpus of english: The penn treebank.
\newblock \emph{Using Large Corpora}, 273.

\bibitem[{Maynez et~al.(2020)Maynez, Narayan, Bohnet, and
  McDonald}]{maynez-etal-2020-faithfulness}
Joshua Maynez, Shashi Narayan, Bernd Bohnet, and Ryan McDonald. 2020.
\newblock \href {https://doi.org/10.18653/v1/2020.acl-main.173} {On
  faithfulness and factuality in abstractive summarization}.
\newblock In \emph{Proceedings of the 58th Annual Meeting of the Association
  for Computational Linguistics}, pages 1906--1919, Online. Association for
  Computational Linguistics.

\bibitem[{McKeown(2020)}]{mckeown-keynote}
Kathleen McKeown. 2020.
\newblock Rewriting the past: Assessing the field through the lens of language
  generation.

\bibitem[{Nan et~al.(2021{\natexlab{a}})Nan, Nallapati, Wang, Nogueira~dos
  Santos, Zhu, Zhang, McKeown, and Xiang}]{nan-etal-2021-entity}
Feng Nan, Ramesh Nallapati, Zhiguo Wang, Cicero Nogueira~dos Santos, Henghui
  Zhu, Dejiao Zhang, Kathleen McKeown, and Bing Xiang. 2021{\natexlab{a}}.
\newblock \href {https://doi.org/10.18653/v1/2021.eacl-main.235} {Entity-level
  factual consistency of abstractive text summarization}.
\newblock In \emph{Proceedings of the 16th Conference of the European Chapter
  of the Association for Computational Linguistics: Main Volume}, pages
  2727--2733, Online. Association for Computational Linguistics.

\bibitem[{Nan et~al.(2021{\natexlab{b}})Nan, Nogueira~dos Santos, Zhu, Ng,
  McKeown, Nallapati, Zhang, Wang, Arnold, and Xiang}]{nan-etal-2021-improving}
Feng Nan, Cicero Nogueira~dos Santos, Henghui Zhu, Patrick Ng, Kathleen
  McKeown, Ramesh Nallapati, Dejiao Zhang, Zhiguo Wang, Andrew~O. Arnold, and
  Bing Xiang. 2021{\natexlab{b}}.
\newblock \href {https://doi.org/10.18653/v1/2021.acl-long.536} {Improving
  factual consistency of abstractive summarization via question answering}.
\newblock In \emph{Proceedings of the 59th Annual Meeting of the Association
  for Computational Linguistics and the 11th International Joint Conference on
  Natural Language Processing (Volume 1: Long Papers)}, pages 6881--6894,
  Online. Association for Computational Linguistics.

\bibitem[{Narayan et~al.(2021)Narayan, Zhao, Maynez, Sim{\~o}es, Nikolaev, and
  McDonald}]{narayan-etal-2021-planning}
Shashi Narayan, Yao Zhao, Joshua Maynez, Gon{\c{c}}alo Sim{\~o}es, Vitaly
  Nikolaev, and Ryan McDonald. 2021.
\newblock \href {https://doi.org/10.1162/tacl_a_00438} {Planning with learned
  entity prompts for abstractive summarization}.
\newblock \emph{Transactions of the Association for Computational Linguistics},
  9:1475--1492.

\bibitem[{Pagnoni et~al.(2021)Pagnoni, Balachandran, and
  Tsvetkov}]{pagnoni-etal-2021-understanding}
Artidoro Pagnoni, Vidhisha Balachandran, and Yulia Tsvetkov. 2021.
\newblock \href {https://doi.org/10.18653/v1/2021.naacl-main.383}
  {Understanding factuality in abstractive summarization with {FRANK}: A
  benchmark for factuality metrics}.
\newblock In \emph{Proceedings of the 2021 Conference of the North American
  Chapter of the Association for Computational Linguistics: Human Language
  Technologies}, pages 4812--4829, Online. Association for Computational
  Linguistics.

\bibitem[{Peyrard and Gurevych(2018)}]{peyrard-gurevych-2018-objective}
Maxime Peyrard and Iryna Gurevych. 2018.
\newblock \href {https://doi.org/10.18653/v1/N18-2103} {Objective function
  learning to match human judgements for optimization-based summarization}.
\newblock In \emph{Proceedings of the 2018 Conference of the North {A}merican
  Chapter of the Association for Computational Linguistics: Human Language
  Technologies, Volume 2 (Short Papers)}, pages 654--660, New Orleans,
  Louisiana. Association for Computational Linguistics.

\bibitem[{Phan et~al.(2021)Phan, Anibal, Tran, Chanana, Bahadroglu, Peltekian,
  and Altan-Bonnet}]{phan2021scifive}
Long~N Phan, James~T Anibal, Hieu Tran, Shaurya Chanana, Erol Bahadroglu, Alec
  Peltekian, and Gr{\'e}goire Altan-Bonnet. 2021.
\newblock Scifive: a text-to-text transformer model for biomedical literature.
\newblock \emph{arXiv preprint arXiv:2106.03598}.

\bibitem[{Pyysalo et~al.(2015)Pyysalo, Ohta, Rak, Rowley, Chun, Jung, Choi,
  Tsujii, and Ananiadou}]{pyysalo2015overview}
Sampo Pyysalo, Tomoko Ohta, Rafal Rak, Andrew Rowley, Hong-Woo Chun, Sung-Jae
  Jung, Sung-Pil Choi, Jun'ichi Tsujii, and Sophia Ananiadou. 2015.
\newblock Overview of the cancer genetics and pathway curation tasks of bionlp
  shared task 2013.
\newblock \emph{BMC bioinformatics}, 16(10):1--19.

\bibitem[{Qi et~al.(2020)Qi, Zhang, Zhang, Bolton, and Manning}]{qi2020stanza}
Peng Qi, Yuhao Zhang, Yuhui Zhang, Jason Bolton, and Christopher~D Manning.
  2020.
\newblock Stanza: A python natural language processing toolkit for many human
  languages.
\newblock \emph{arXiv preprint arXiv:2003.07082}.

\bibitem[{Raffel et~al.(2020)Raffel, Shazeer, Roberts, Lee, Narang, Matena,
  Zhou, Li, Liu et~al.}]{raffel2020exploring}
Colin Raffel, Noam Shazeer, Adam Roberts, Katherine Lee, Sharan Narang, Michael
  Matena, Yanqi Zhou, Wei Li, Peter~J Liu, et~al. 2020.
\newblock Exploring the limits of transfer learning with a unified text-to-text
  transformer.
\newblock \emph{J. Mach. Learn. Res.}, 21(140):1--67.

\bibitem[{Sun et~al.(2019)Sun, Shapira, Dagan, and
  Nenkova}]{sun-etal-2019-compare}
Simeng Sun, Ori Shapira, Ido Dagan, and Ani Nenkova. 2019.
\newblock \href {https://doi.org/10.18653/v1/W19-2303} {How to compare
  summarizers without target length? pitfalls, solutions and re-examination of
  the neural summarization literature}.
\newblock In \emph{Proceedings of the Workshop on Methods for Optimizing and
  Evaluating Neural Language Generation}, pages 21--29, Minneapolis, Minnesota.
  Association for Computational Linguistics.

\bibitem[{Tang et~al.(2022)Tang, Nair, Wang, Wang, Desai, Wade, Li,
  Celikyilmaz, Mehdad, and Radev}]{tang-etal-2022-confit}
Xiangru Tang, Arjun Nair, Borui Wang, Bingyao Wang, Jai Desai, Aaron Wade,
  Haoran Li, Asli Celikyilmaz, Yashar Mehdad, and Dragomir Radev. 2022.
\newblock \href {https://doi.org/10.18653/v1/2022.naacl-main.415} {{CONFIT}:
  Toward faithful dialogue summarization with linguistically-informed
  contrastive fine-tuning}.
\newblock In \emph{Proceedings of the 2022 Conference of the North American
  Chapter of the Association for Computational Linguistics: Human Language
  Technologies}, pages 5657--5668, Seattle, United States. Association for
  Computational Linguistics.

\bibitem[{Uzuner et~al.(2011)Uzuner, South, Shen, and DuVall}]{uzuner20112010}
{\"O}zlem Uzuner, Brett~R South, Shuying Shen, and Scott~L DuVall. 2011.
\newblock 2010 i2b2/va challenge on concepts, assertions, and relations in
  clinical text.
\newblock \emph{Journal of the American Medical Informatics Association},
  18(5):552--556.

\bibitem[{Vijayakumar et~al.(2016)Vijayakumar, Cogswell, Selvaraju, Sun, Lee,
  Crandall, and Batra}]{vijayakumar2016diverse}
Ashwin~K Vijayakumar, Michael Cogswell, Ramprasath~R Selvaraju, Qing Sun,
  Stefan Lee, David Crandall, and Dhruv Batra. 2016.
\newblock Diverse beam search: Decoding diverse solutions from neural sequence
  models.
\newblock \emph{arXiv preprint arXiv:1610.02424}.

\bibitem[{Wadden et~al.(2020)Wadden, Lin, Lo, Wang, van Zuylen, Cohan, and
  Hajishirzi}]{wadden-etal-2020-fact}
David Wadden, Shanchuan Lin, Kyle Lo, Lucy~Lu Wang, Madeleine van Zuylen, Arman
  Cohan, and Hannaneh Hajishirzi. 2020.
\newblock \href {https://doi.org/10.18653/v1/2020.emnlp-main.609} {Fact or
  fiction: Verifying scientific claims}.
\newblock In \emph{Proceedings of the 2020 Conference on Empirical Methods in
  Natural Language Processing (EMNLP)}, pages 7534--7550, Online. Association
  for Computational Linguistics.

\bibitem[{Wadden et~al.(2022)Wadden, Lo, Wang, Cohan, Beltagy, and
  Hajishirzi}]{wadden-etal-2022-multivers}
David Wadden, Kyle Lo, Lucy Wang, Arman Cohan, Iz~Beltagy, and Hannaneh
  Hajishirzi. 2022.
\newblock \href {https://doi.org/10.18653/v1/2022.findings-naacl.6}
  {{M}ulti{V}er{S}: Improving scientific claim verification with weak
  supervision and full-document context}.
\newblock In \emph{Findings of the Association for Computational Linguistics:
  NAACL 2022}, pages 61--76, Seattle, United States. Association for
  Computational Linguistics.

\bibitem[{Wan and Bansal(2022)}]{wan-bansal-2022-factpegasus}
David Wan and Mohit Bansal. 2022.
\newblock \href {https://doi.org/10.18653/v1/2022.naacl-main.74}
  {{F}act{PEGASUS}: Factuality-aware pre-training and fine-tuning for
  abstractive summarization}.
\newblock In \emph{Proceedings of the 2022 Conference of the North American
  Chapter of the Association for Computational Linguistics: Human Language
  Technologies}, pages 1010--1028, Seattle, United States. Association for
  Computational Linguistics.

\bibitem[{Wieting and Gimpel(2017)}]{wieting2017paranmt}
John Wieting and Kevin Gimpel. 2017.
\newblock Paranmt-50m: Pushing the limits of paraphrastic sentence embeddings
  with millions of machine translations.
\newblock \emph{arXiv preprint arXiv:1711.05732}.

\bibitem[{Wolf et~al.(2020)Wolf, Debut, Sanh, Chaumond, Delangue, Moi, Cistac,
  Rault, Louf, Funtowicz, Davison, Shleifer, von Platen, Ma, Jernite, Plu, Xu,
  Le~Scao, Gugger, Drame, Lhoest, and Rush}]{transformers}
Thomas Wolf, Lysandre Debut, Victor Sanh, Julien Chaumond, Clement Delangue,
  Anthony Moi, Pierric Cistac, Tim Rault, Remi Louf, Morgan Funtowicz, Joe
  Davison, Sam Shleifer, Patrick von Platen, Clara Ma, Yacine Jernite, Julien
  Plu, Canwen Xu, Teven Le~Scao, Sylvain Gugger, Mariama Drame, Quentin Lhoest,
  and Alexander Rush. 2020.
\newblock \href {https://doi.org/10.18653/v1/2020.emnlp-demos.6} {Transformers:
  State-of-the-art natural language processing}.
\newblock In \emph{Proceedings of the 2020 Conference on Empirical Methods in
  Natural Language Processing: System Demonstrations}, pages 38--45, Online.
  Association for Computational Linguistics.

\bibitem[{Wu et~al.(2020)Wu, Ma, Wu, Manyumwa, and
  Ji}]{wu-etal-2020-unsupervised}
Hanlu Wu, Tengfei Ma, Lingfei Wu, Tariro Manyumwa, and Shouling Ji. 2020.
\newblock \href {https://doi.org/10.18653/v1/2020.emnlp-main.294} {Unsupervised
  reference-free summary quality evaluation via contrastive learning}.
\newblock In \emph{Proceedings of the 2020 Conference on Empirical Methods in
  Natural Language Processing (EMNLP)}, pages 3612--3621, Online. Association
  for Computational Linguistics.

\bibitem[{Xiao et~al.(2022)Xiao, Beltagy, Carenini, and Cohan}]{primera}
Wen Xiao, Iz~Beltagy, Giuseppe Carenini, and Arman Cohan. 2022.
\newblock \href {https://doi.org/10.18653/v1/2022.acl-long.360} {{PRIMERA}:
  Pyramid-based masked sentence pre-training for multi-document summarization}.
\newblock In \emph{Proceedings of the 60th Annual Meeting of the Association
  for Computational Linguistics (Volume 1: Long Papers)}, pages 5245--5263,
  Dublin, Ireland. Association for Computational Linguistics.

\bibitem[{Yuan et~al.(2021)Yuan, Neubig, and Liu}]{yuan2021BARTScore}
Weizhe Yuan, Graham Neubig, and Pengfei Liu. 2021.
\newblock Bartscore: Evaluating generated text as text generation.
\newblock \emph{Advances in Neural Information Processing Systems},
  34:27263--27277.

\bibitem[{Zhang et~al.(2020)Zhang, Zhao, Saleh, and Liu}]{zhang2020pegasus}
Jingqing Zhang, Yao Zhao, Mohammad Saleh, and Peter Liu. 2020.
\newblock Pegasus: Pre-training with extracted gap-sentences for abstractive
  summarization.
\newblock In \emph{International Conference on Machine Learning}, pages
  11328--11339. PMLR.

\bibitem[{Zhang et~al.(2022)Zhang, Wan, and Bansal}]{zhang2022extractive}
Shiyue Zhang, David Wan, and Mohit Bansal. 2022.
\newblock \href {http://arxiv.org/abs/2209.03549} {Extractive is not faithful:
  An investigation of broad unfaithfulness problems in extractive
  summarization}.

\bibitem[{Zhang et~al.(2019{\natexlab{a}})Zhang, Kishore, Wu, Weinberger, and
  Artzi}]{zhang2019BERTScore}
Tianyi Zhang, Varsha Kishore, Felix Wu, Kilian~Q Weinberger, and Yoav Artzi.
  2019{\natexlab{a}}.
\newblock Bertscore: Evaluating text generation with bert.
\newblock \emph{arXiv preprint arXiv:1904.09675}.

\bibitem[{Zhang et~al.(2019{\natexlab{b}})Zhang, Baldridge, and
  He}]{zhang-etal-2019-paws}
Yuan Zhang, Jason Baldridge, and Luheng He. 2019{\natexlab{b}}.
\newblock \href {https://doi.org/10.18653/v1/N19-1131} {{PAWS}: Paraphrase
  adversaries from word scrambling}.
\newblock In \emph{Proceedings of the 2019 Conference of the North {A}merican
  Chapter of the Association for Computational Linguistics: Human Language
  Technologies, Volume 1 (Long and Short Papers)}, pages 1298--1308,
  Minneapolis, Minnesota. Association for Computational Linguistics.

\bibitem[{Zhang et~al.(2021)Zhang, Zhang, Qi, Manning, and
  Langlotz}]{zhang2021biomedical}
Yuhao Zhang, Yuhui Zhang, Peng Qi, Christopher~D Manning, and Curtis~P
  Langlotz. 2021.
\newblock Biomedical and clinical {E}nglish model packages for the {S}tanza
  {P}ython {NLP} library.
\newblock \emph{Journal of the American Medical Informatics Association}.

\bibitem[{Zhao et~al.(2022)Zhao, Khalman, Joshi, Narayan, Saleh, and
  Liu}]{zhao2022calibrating}
Yao Zhao, Misha Khalman, Rishabh Joshi, Shashi Narayan, Mohammad Saleh, and
  Peter~J Liu. 2022.
\newblock Calibrating sequence likelihood improves conditional language
  generation.
\newblock \emph{arXiv preprint arXiv:2210.00045}.

\bibitem[{Zhou et~al.(2021)Zhou, Neubig, Gu, Diab, Guzm{\'a}n, Zettlemoyer, and
  Ghazvininejad}]{zhou-etal-2021-detecting}
Chunting Zhou, Graham Neubig, Jiatao Gu, Mona Diab, Francisco Guzm{\'a}n, Luke
  Zettlemoyer, and Marjan Ghazvininejad. 2021.
\newblock \href {https://doi.org/10.18653/v1/2021.findings-acl.120} {Detecting
  hallucinated content in conditional neural sequence generation}.
\newblock In \emph{Findings of the Association for Computational Linguistics:
  ACL-IJCNLP 2021}, pages 1393--1404, Online. Association for Computational
  Linguistics.

\bibitem[{Zhou and Bhat(2021)}]{zhou-bhat-2021-paraphrase}
Jianing Zhou and Suma Bhat. 2021.
\newblock \href {https://doi.org/10.18653/v1/2021.emnlp-main.414} {Paraphrase
  generation: A survey of the state of the art}.
\newblock In \emph{Proceedings of the 2021 Conference on Empirical Methods in
  Natural Language Processing}, pages 5075--5086, Online and Punta Cana,
  Dominican Republic. Association for Computational Linguistics.

\bibitem[{Zhu et~al.(2021)Zhu, Hinthorn, Xu, Zeng, Zeng, Huang, and
  Jiang}]{zhu-etal-2021-enhancing}
Chenguang Zhu, William Hinthorn, Ruochen Xu, Qingkai Zeng, Michael Zeng,
  Xuedong Huang, and Meng Jiang. 2021.
\newblock \href {https://doi.org/10.18653/v1/2021.naacl-main.58} {Enhancing
  factual consistency of abstractive summarization}.
\newblock In \emph{Proceedings of the 2021 Conference of the North American
  Chapter of the Association for Computational Linguistics: Human Language
  Technologies}, pages 718--733, Online. Association for Computational
  Linguistics.

\bibitem[{Zhu et~al.(2018)Zhu, Lu, Zheng, Guo, Zhang, Wang, and
  Yu}]{zhu2018texygen}
Yaoming Zhu, Sidi Lu, Lei Zheng, Jiaxian Guo, Weinan Zhang, Jun Wang, and Yong
  Yu. 2018.
\newblock Texygen: A benchmarking platform for text generation models.
\newblock In \emph{The 41st International ACM SIGIR Conference on Research \&
  Development in Information Retrieval}, pages 1097--1100.

\end{thebibliography}
\bibliographystyle{acl_natbib}

\appendix

\section{Clinical Dataset} \label{app:clinical-dataset}

 As in \citet{adams-etal-2021-whats}, references are extracted from the Brief Hospital Course section of discharge summaries from the publicly-available MIMIC-III dataset \citep{johnson2016mimic}, and the source text consists of all available notes written between admission and discharge regarding a single patient. It is a highly noisy, naturally occurring dataset, which we expect to present challenges for faithfulness.

\section{Negative Methods} \label{app:neg-methods}

\paragraph{Negative Methods.} \textbf{Mask-And-Fill} involves masking portions of a reference summary, and using a pre-trained language model to fill in the blanks. It has been used for contrastive fine-tuning \citep{cliff}, evaluation \citep{ctc}, and fine-grained optimization of noisy references \citep{zhou-etal-2021-detecting}. First, following \citet{goyal-durrett-2021-annotating, lee-etal-2022-masked}, we identify all noun phrases\footnote{`NP' using the annotation scheme from the Penn Treebank \citep{marcinkiewicz1994building}.} as candidates for masking using Stanza's constituency parser \citep{qi2020stanza}. Then, we sample a subset of non overlapping phrases to mask and generate replacements with SciFive \citep{phan2021scifive}. SciFive is a language model pre-trained on diverse biomedical tasks with T5-inspired \citep{raffel2020exploring} prefixes. We perform a beam search of size 4 to generate in-filled text for each spans and set the minimum generated tokens to be equal to the number of masked tokens to preserve length. \linebreak

\noindent\textit{Hyper-Parameters of Significance}: the target token mask rate: $m$, which defines the percentage of noun phrases from the unmasked reference to mask. We vary $m$ to measure the impact of corruption `intensity' on the efficacy of contrastive fine-tuning. \linebreak

\noindent For \textbf{Entity swapping} \citep{factcc}, we replace reference entities and numbers with entities and numbers from the source text (\texttt{intrinsic} hallucinations) or the corpus (\texttt{extrinsic}). Please refer to Appendix \ref{app:neg-methods} for more details. \linebreak

\noindent\textit{Hyper-Parameters of Significance}: the swap rate: $s$, which defines the percentage of named entities and numbers in the reference, separately, to replace.

Entity and number swapping was initially proposed for faithfulness evaluation (FactCC \citep{factcc}) and has subsequently been used for contrastive fine-tuning \citep{tang-etal-2022-confit} and post-hoc editing \citep{cao-etal-2020-factual, chen-etal-2021-improving, zhu-etal-2021-enhancing}, etc. For each corpora, we extract numbers with numbers with \href{https://github.com/nielstron/quantulum3}{quantulum3}. Separately for each corpora, we extract named entities relevant to each domain. For chemistry, we extract chemicals and other types\footnote{The list of types includes genes, diseases, species, mutations, cell lines, and cell types.} with BERN2 \citep{kim2019neural}. BERN2 is trained on PubMed articles to identify chemicals and diseases and link them to a unique identifier (CUI) in the Unified Medical Language System (UMLS) \citep{bodenreider2004unified}. For the clinical corpus, we use the Stanza transformer model \citep{qi2020stanza, zhang2021biomedical} trained on the i2b2 corpus \citep{uzuner20112010}, which learns to identify patient problems, tests, and treatments. Finally, for biomedical, we use the Stanza model trained on the BioNLP13CG corpus \citep{pyysalo2015overview}, which includes a diverse set of 13 categories.

To simulate intrinsic errors, we perform swaps at random with entities of the same semantic category from the source document. For extrinsic, we also restrict the swap to be from the same semantic category, yet sample from the entire corpus.

\begin{figure*}[t]
\centering
\includegraphics[width=\linewidth]{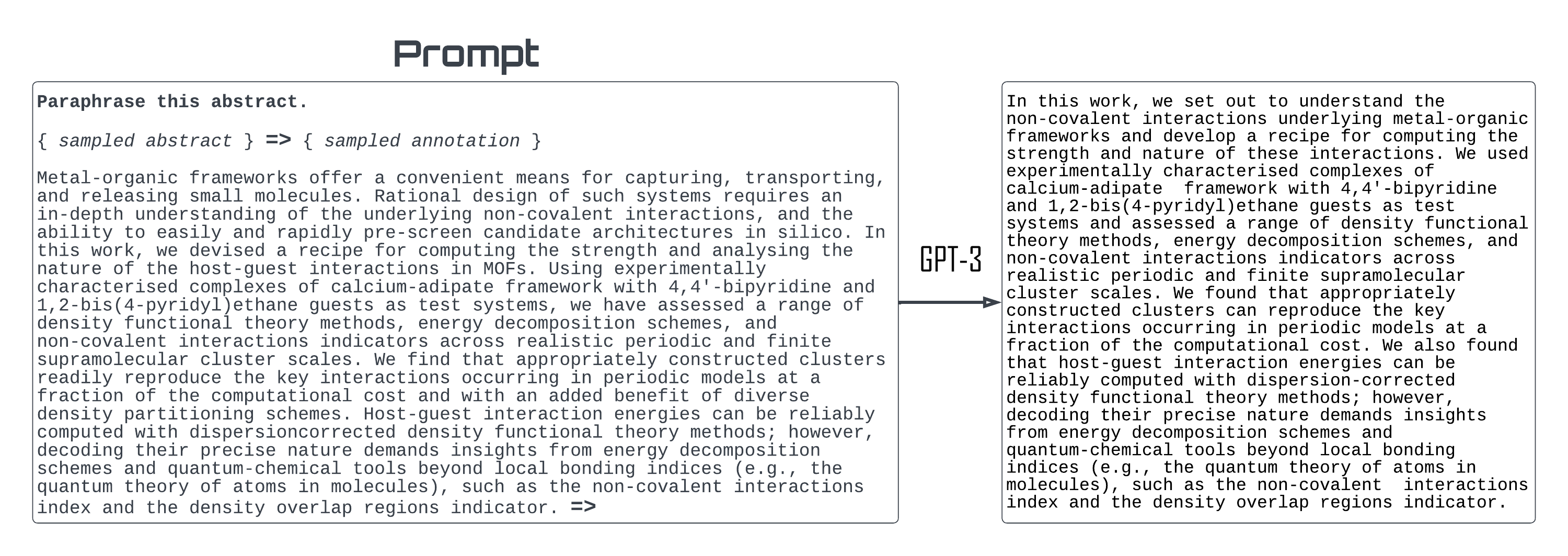}
\caption{An example prompt and paraphrase output from GPT-3. Words are changed but the meaning is preserved. } 
\label{fig:gpt3-paraphrase}
\end{figure*}

\section{GPT-3 as a Paraphraser} \label{app:gpt3}

Paraphrasing is typically done with synonym substitution \citep{zhou-bhat-2021-paraphrase}, neural models \citep{goyal-durrett-2020-neural} trained on paraphrase corpora \citep{wieting2017paranmt, zhang-etal-2019-paws}, or back-translation \citep{factcc, fabbri-etal-2021-improving}. Yet, these methods performed very poorly on our long scientific texts, likely due to highly specialized lexicons and lack of large-scale, domain-specific paraphrase corpora. In Figure \ref{fig:gpt3-paraphrase}, we show an example prompt and sampled paraphrase from one-shot paraphrasing with GPT-3. A random sample of one annotation pair, as well as the abstract to be paraphrased, are then provided as prompts, which are both preceeded by a fixed instruction: \texttt{Paraphrase this abstract.} for abstract generation, and \texttt{Paraphrase this Summary.} for clinical summarization). We sample 1 due to token limits yet prompt sampling also increases diversity, as shown in \citet{chintagunta-etal-2021-medically}.

A softmax temperature $t$ of 0.7 is used to sample 5 unique outputs from GPT-3 (\texttt{text-davinci-002}).

\begin{table*}[t]
\centering
\small
\begin{tabular}{cl|ccc|ccc|ccc}
\hline
& \multirow{2}{*}{\texttt{\makecell{Candidate \\ Method}}} & \multicolumn{3}{c}{\texttt{Clinical}} & \multicolumn{3}{c}{\texttt{Chemical}} & \multicolumn{3}{c}{\texttt{Biomedical}} \\
& & Rel. & Faith. & Extract. & Rel. & Faith. & Extract. & Rel. & Faith. & Extract. \\
\hline
\multirow{8}{*}{\makecell{Faith. \\ Contrast}} & \textcolor{red}{Mask-And-Fill} (\emph{Low}) & 0.98 & 0.52 & 1.55 & 0.99 & 0.75 & 3.24 & 0.97 & 0.73 & 4.92 \\
& \textcolor{red}{Mask-And-Fill} (\emph{High}) & 0.97 & 0.52 & 1.44 & 0.97 & 0.73 & 2.90 & 0.95 & 0.71 & 4.05 \\
& \textcolor{red}{Swap Intrinsic} (\emph{Low}) & 0.94 & 0.52 & 1.64 & 0.97 & 0.70 & 2.92 & 0.98 & 0.71 & 4.70 \\
& \textcolor{red}{Swap Intrinsic} (\emph{High}) & 0.90 & 0.52 & 1.82 & 0.95 & 0.65 & 2.62 & 0.97 & 0.67 & 4.13 \\
& \textcolor{red}{Swap Extrinsic} (\emph{Low}) & 0.94 & 0.52 & 1.64 & 0.97 & 0.70 & 2.92 & 0.98 & 0.68 & 4.44 \\
& \textcolor{red}{Swap Extrinsic} (\emph{High}) & 0.90 & 0.52 & 1.82 & 0.95 & 0.65 & 2.62 & 0.97 & 0.64 & 3.79 \\ 
& \textcolor{Green}{Paraphrase} & 0.90 & 0.52 & 1.26 & 0.94 & 0.77 & 3.06 & 0.92 & 0.73 & 4.00 \\
& \textcolor{Green}{Reference} & 1.00 & 0.52 & 1.96 & 1.00 & 0.76 & 3.54 & 1.00 & 0.74 & 5.78\\ \hline
\multirow{2}{*}{\makecell{Rel. \\ Rank}} & Diverse Beam (PRIMERA) & 0.84 & 0.53 & 2.65 & 0.87 & 0.85 & 9.66 & 0.86 & 0.86 & 12.90 \\
& Diverse Beam (LongT5) & 0.83 & 0.52 & 2.06 & 0.86 & 0.83 & 7.46 & 0.85 & 0.82 & 8.39 \\ \hline
\end{tabular}
\caption{Statistics for each candidate generation method. Rel. stands for Relevance and is measured by BERTScore F1 overlap with the reference. Faith. stands for faithfulness and is measured by the FactScore (as defined in \S \ref{sec:faithful-setup}). Extract. stands for the extractive density (level of copy-and-paste) as defined by \citet{grusky2018newsroom}. The first 6 rows (Mask-And-Fill and Swaps) construct negative examples for faithfulness calibration. The next two rows form the positive candidate set for faithfulness. The last two (diverse beam) form candidates for relevance calibration.} \label{tab:method-metrics}
\end{table*}

\section{Evaluation Metrics} \label{app:metrics}

\subsection{Relevance}

For BERTScore \citep{zhang2019BERTScore}, we use \textit{allenai/scibert\_scivocab\_uncased} weights and all default settings from HuggingFace \citep{transformers}. We normalize by subtracting each metric by its mean and then dividing by the standard deviation to account for metrics with different scales. We use test set fine-tuning (\texttt{FT}) scores to compute mean and standard deviation so that $Rel_{Agg}$ is $0$ after \texttt{FT} and $>0$ values are standard deviation improvements from calibration.

\subsection{Faithfulness}

For BARTScore, we use a PEGASUS \citep{zhang2020pegasus} model pretrained on the PubMed summarization corpus\footnote{\texttt{google/pegasus-pubmed } on the HuggingFace Transformers Hub \citep{transformers}.} for the PubMed and Clinical datsets, and we use a Longformer Encoder-Decoder \citep{beltagy2020longformer} trained on a more faithful, synthetic version of our clinical corpus from \citet{adams2022learning}. We report the average log-likelihood of each candidate summary $S$: $\frac{1}{|S|} \sum_{i \in |S|}{p(s_i|, s_{j < i}, D)}$. BARTScore and BERTScore are not explicitly trained to detect domain-specific errors. As such, we implement \textbf{FactScore}, which is based on the state of the art model (MultiVERS \citep{wadden-etal-2022-multivers}) trained on the SciFact scientific claims dataset \citep{wadden-etal-2020-fact}. SciFact is an expert-annotated dataset of 1,409 sentence-level scientific claims. We first align each summary sentence to a handful of sentences (1-5) from the source document, following the greedy algorithm from \citet{lebanoff-etal-2019-analyzing}. Then we score each sentence based on its alignment and average the \texttt{SUPPORTED} label prediction probabilities.

\section{Candidate Set Analysis (Ctd.)} \label{app:set-analysis}

The idea behind generating candidates with different methods and parameters is twofold: (1) to better understand which candidate generation methods work best on our task of interest: long-form scientific summarization, and (2) to end up with a diverse candidate pool, which allows us to effectively control for certain characteristics when selecting final subsets for calibration experiments.

In Table \ref{tab:method-metrics}, we show statistics (relevance, faithfulness, and extractive density) for each candidate generation method across the three datasets.

\paragraph{Analysis.} As noted in \citet{adams2022learning}, the references for the clinical dataset are very abstractive (1.96 density) and unfaithful (0.52 FactScore), as compared to the chemical (3.54 / 0.76) and biomedical (5.78 / 0.74) data. The former is affected by missing clinical notes while the latter references are abstracts, which \emph{should} be mostly entailed by the claims made in the main paper. Interestingly, the reference is deemed less faithful than the model generations (0.52 vs 0.53/0.52, 0.76 vs 0.85/0.83, and 0.74 vs 0.86/0.82 for diverse beam search clinical, chemical, and biomedical). This likely has to do with the fact that the fine-tuned models (PRIMERA and LongT5) perform substantially more copy-and-pasting from the source input as the references (1.96 vs 2.65/2.06, 3.54 vs 9.66/7.46, and 5.78 vs 12.90/8.39, respectively).

The most unfaithful corruption method is Swap. When looking at (High) across Intrinsic and Extrinsic, its FactScores are 0.52/0.52, 0.65/0.65, and 0.67/0.64 versus 0.52, 0.73, 0.71 for Mask-And-Fill (High), respectively. This likely has to do with an in-domain LM (SciFive) making reasonably well-informed replacements for noun phrases, whereas entity swapping is indiscriminate and random. The (High) parameter settings for Mask-And-Fill and Swap create less faithful candidates vis-a-vis the (Low) settings (0.75/0.70/0.70 versus 0.73/0.65/0.65 for High and Low on Chemical, for example), as expected. Replacing more text from the references introduces more factual errors.

The PRIMERA model produces more extractive summaries with diverse beam search (2.65/9.66/12.90 vs 2.06/7.46/8.39), which are scored as more relevant and faithful than LongT5.

\section{Training Details} \label{app:training-details}

\subsection{FT Training Details}

We fine-tune (\texttt{FT}) two state of the art long-document summarization models for 50,000 steps: PRIMERA \citep{primera} (the backbone is a Longformer Encoder-Decoder (LED) \citep{beltagy2020longformer} model) and LongT5 \citep{longt5} (which incorporates the sparse attention of ETC \citep{ainslie-etal-2020-etc} into PEGASUS \citep{zhang2020pegasus}) on a single A100 40GB GPU with half precision (FP16)\footnote{Only for PRIMERA since LongT5 does not support half precision weights.}) and a batch a size of 1 (with 16 gradient accumulation steps). We set the  maximum learning rate to $3e-5$ with 2,000 warmup steps, followed by a linear decay. We set a maximum input sequence length of 4,096 for both models\footnote{Even though LongT5 has a maximum input sequence length of 16,384, we chose 4,096 to match PRIMERA and because of GPU memory constraints.}, and a maximum target length of 512 for training / inference for abstract generation (Chemical and Biomedical) and 256 for clinical summarization. Each fine-tuning (\texttt{FT}) experiment took $\sim3.5$ days.

We select the better performing model (PRIMERA) as the model to be used for \texttt{CT} (See Table \ref{tab:ft-results}). As discussed in \S \ref{sec:relevance-setup}, LongT5 is still used to supply ten diverse summaries to the candidate pool for relevance calibration.

\begin{table}[h]
\centering
\small
\begin{tabular}{cl|ccc}
& \textbf{Parameter} & \textbf{Clin} & \textbf{Chem} & \textbf{Bio} \\ \hline
\multirow{5}{*}{\makecell{Relevance \\ Ranking}} & $\lambda_{MLE}$ & 0.1 & 0.1 & 0.1  \\
 & $\lambda_{CA}$ & 1.0 & 1.0 & 1.0  \\
 & $\lambda_{margin}$ & .001 & .001 & .001  \\
 & $\alpha$ (length penalty) & 1.0 & 2.0 & 2.0  \\
 & $\tau$ (scale) & .01 & 0.1 & 0.1  \\ \hline
 \multirow{2}{*}{\makecell{Faithful \\ Contrast}} & $\lambda_{MLE}$ & 1.0 & 1.0 & 1.0  \\
 & $\lambda_{CA}$ & 1.0 & 10.0 & 1.0  \\ \hline
\end{tabular}
\caption{Hyper-Parameters for calibration fine-tuning.} \label{tab:hparams}
\end{table}

\subsection{CT Training Details}

We run calibration-tuning (\texttt{CT}) for a maximum of 10,000 steps and select the checkpoint which maximizes either $Rel_{Agg}$ or $Faith_{Agg}$ (depending on the experiment) on the validation set in 1,000 step intervals.

We use the same hyper-parameters as $FT$ except the batch size is reduced from 16 to 8. Hyper-parameters related to the \texttt{CT} loss function were tuned separately for each dataset and quality metric (the values selected are shown in Table \ref{tab:hparams}). Each \texttt{CT} experiment took $\sim1$ day to train.

As in \citet{longt5}, summaries are generated greedily, which we found to be significantly faster and even slightly outperformed beam search\footnote{This also means that a length penalty cannot be applied during decoding, which puts more emphasis on the significant role of length tuning during relevance calibration.}.

\section{Identifying Possible Correlates} \label{app:hypothesis}

We examine five basic aspects of calibration sets that \emph{should} have some impact on downstream performance. For each aspect, we provide intuition and some related work to guess the nature of the impact, which we investigate empirically in \S \ref{sec:results}.

\subsection{Overall Quality}

\paragraph{Definition.} For the purposes of this analysis, for relevance-rank sets, we define quality as the average $Rel_{Agg}$ score of the candidates.

\paragraph{Relevance Hypothesis.} For relevance, high-quality sets might be preferable to lower-quality sets for two reasons: (1) the model before calibration (pre-\texttt{CT}) has already been fine-tuned (post-\texttt{FT}) on the same training data used for \texttt{CT}, so it likely already assigns a high-probability mass to summaries which are close to the reference. Candidate summaries which deviate too much should already have a low probability of being generated and thus not provide much of a learning signal. In some ways, this hypothesis is supported by \citet{zhao2022calibrating} who find that using a model's top beams produces consistently better results than diverse beam search or sampling-based methods (e.g., nucleus sampling \citep{holtzman2019curious}). There is an inherent tension between the calibration objective, which involves exploration, and the MLE, which assigns all probability mass to a single point.

\subsection{Margin}

Overall quality covers average metric values, while margin covers within-set variation in quality.

\paragraph{Definition.} For relevance rank-based sets, we define the margin as the average relevance score between all adjacent pairs of ranked candidates: $Avg(Rel_{Agg}(\hat{S_i}, S) - Rel_{Agg}(\hat{S_{i+1}}, S)), i \in |\bm{\hat{S}}| - 1$. For faithfulness, we define it as the delta in average $Faith_{Agg}$ scores for summaries in 
 the positive and negative contrast sets, respectively.

\paragraph{Relevance Hypothesis.} As noisy proxies for human judgments \citep{peyrard-gurevych-2018-objective}, subtle differences in relevance metrics (e.g, ROUGE and BERTScore) might not be meaningful. As such, we hypothesize that, all else equal, sets with larger metric gaps will provide a clearer training signal during calibration and superior downstream results.

\paragraph{Faithfulness Hypothesis.} Trivially, one would want positive candidates which are fully faithful. For negatives, it is less clear. The emphasis in the literature has been on producing negative summaries which mimic model errors \citep{goyal-durrett-2021-annotating}. Yet, less is discussed about the intensity of errors. \citet{lee-etal-2022-masked} explore corruption intensity in the context of training a faithfulness evaluator, and the results suggest a concave relationship. Too few edits and the contrast sets are not easily separable, yet too dramatic, and the contrastive loss is ineffectual. We suspect a similar result for calibrating with a contrastive objective.

\subsection{Lexical Diversity}

The previous calibration set characteristic (Margin) covered metric-based comparisons. In this section, we perform comparisons solely at the word-level.

\paragraph{Definition.} We define lexical diversity as the average pairwise self-BLEU score \citep{zhu2018texygen, alihosseini-etal-2019-jointly} between all candidates in a relevance ranking set and separately, for positives and negative subsets in a faithfulness contrast set. 

\paragraph{Relevance Hypothesis.} All else equal, high lexical diversity should improve the robustness of calibration models as it somewhat dampens some of the noise from single-reference MLE training\footnote{We use the word \emph{somewhat} because we acknowledge that relevance metrics measure overlap to a single reference, so introducing diverse calibration candidates does not necessarily encourage, or reward, more diverse outputs. Access to multiple references, or calibrating against human judgments, would better mitigate the single reference exposure bias problem.}.

\paragraph{Faithfulness Hypothesis.} High lexical diversity within positive and negative sets should make the contrastive classifier less reliant on lexical overlap and focus more on the gap in faithfulness between positive and negatives. Lexical diversity likely means more coverage of error types, which has been shown to be beneficial for contrastive fine-tuning \citep{cao-wang-2021-cliff, adams2022learning}.

\subsection{Likelihood}

This section covers a model-specific aspect of calibration sets: the likelihood of the candidate summaries under the model post-\texttt{FT} and pre-\texttt{CT}.

\paragraph{Definition.} For each candidate summary, we compute its length-normalized conditional log likelihood: $\frac{1}{L}\sum_{l=1}^{L}{log\textit{p}(s_{l}|D, S_{< l}; \theta_{FT})}$, where $\theta_{FT}$ denotes the model parameters after fine-tuning.

\paragraph{Relevance Hypothesis.} One would suspect that likely calibration sets are preferable to unlikely since there is little need to calibrate a model to candidate summaries it was never likely to generate.

\paragraph{Faithfulness Hypothesis.} In a similar vein, it makes sense that contrastive learning for faithulness will be most powerful when the model is most surprised. That is, the negatives are more likely to be generated than the positive. This relates to work by \citet{goyal-durrett-2021-annotating}, who argue that negative sets should mimic observed errors.

\subsection{Spurious Correlates}

Automatic evaluation metrics have a tendency to reward outputs with characteristics which are spuriously correlated to quality \citep{durmus-etal-2022-spurious}.

\paragraph{Definitions.} While many possibilities exist \citep{durmus-etal-2022-spurious}, 
for relevance, we focus on summary length, as defined by number of tokens. For faithfulness, we focus on extractiveness, which we measure with density \citep{grusky2018newsroom}: the average squared length of extractive fragments. It approximates the level of copy-and-paste.

\paragraph{Relevance Hypothesis.} \citet{sun-etal-2019-compare} discover that ROUGE rewards longer summaries while humans prefer concise summaries. We hypothesize that exposing models to longer outputs during calibration will lead to longer summaries, which will have higher relevance scores. By controlling for calibration set length, we can better understand whether or not some of the gains from calibration simply come from length tuning\footnote{While length can be influenced during beam search with minimum/maximum length restrictions and length penalties, these measures do not expose a model to long summaries. }.

\paragraph{Faithfulness Hypothesis.} \citet{ladhak-etal-2022-faithful} note that faithfulness metrics tend to prefer summaries with high levels of extraction, all else equal. Yet, \citet{zhang2022extractive} demonstrate that highly extractive does not always mean more faithful, so it is important to get a sense of how much faithfulness calibration is driven by more copy-and-paste.

\section{Analysis of Spurious Correlates} \label{app:spurious}

\begin{figure}[h]
\centering
\includegraphics[width=\linewidth]{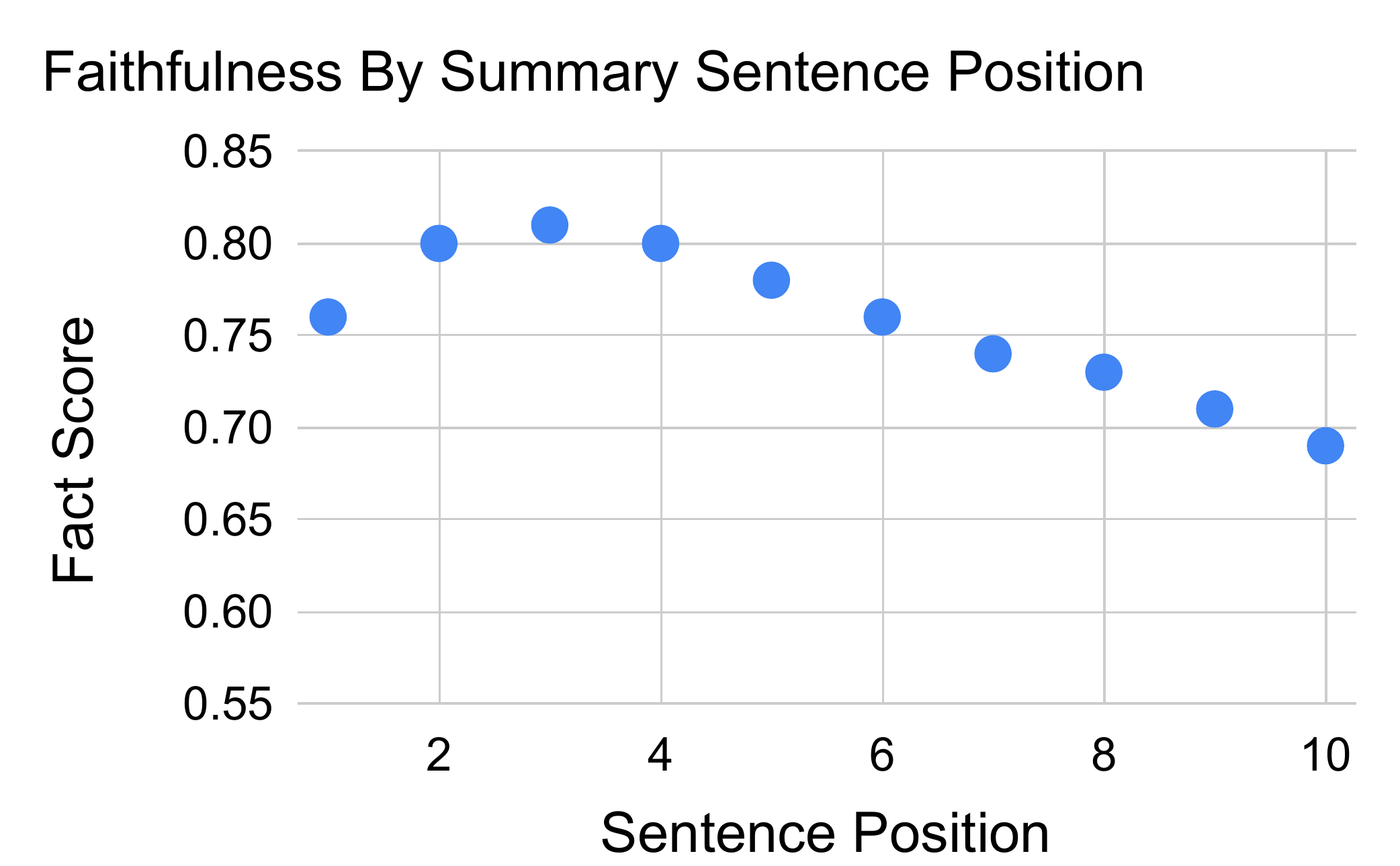}
\caption{Sentence-level faithfulness, as defined by FactScore in \S \ref{sec:faithful-setup}, declines as summaries grow longer.} 
\label{fig:curious-case}
\end{figure}

\subsection{The Outsized Role of Length} \label{app:length}

\paragraph{tl;dr.} The length of summaries is correlated with performance for both relevance and faithful calibration yet for different reasons. For relevance, it can help reduce discrepancies in token-level length between references and generated summaries after fine-tuning. For faithfulness, generated summaries become less faithful as average length increases.

\paragraph{Evidence.} For relevance calibration, the Table~\ref{tab:relevance-results} section on \texttt{Spurious Correlates} shows that selecting the longest summaries is preferable to the shortest for Clinical calibration (.255 versus .181) yet the reverse is true for Biomedical (.017 for max length and .033 for min length). We can trace this to a gap, after fine-tuning, in model summary length and reference lengths. On average, PRIMERA summaries after \texttt{FT} are 119 tokens for clinical and 230 for biomedical. Yet, the clinical references are, on average, 416 tokens and only 205 for biomedical. The optimal length strategy seems contingent on the direction of the length gap.

For faithfulness, we simply compute the correlation between $Faith_{Agg}$ and summary tokens: $-.75$. For faithfulness, we can confirm the presence of text degeneration \citep{holtzman2019curious} as a function of output length by measuring the average $FactScore$ at each sentence position in the summary. Figure \ref{fig:curious-case} confirms this story, despite an initial slight increase up to the third sentence.

\paragraph{Implications.} For relevance, as argued by \citet{sun-etal-2019-compare}, work should acknowledges changes in the lengths of summaries and address its role in impacting relevance metrics. Long-form summarization research which involves identifying and solving subproblems \citep{krishna-etal-2021-generating} might mitigate some of the length-based degeneration.

\begin{table}[h]
\centering
\small
\begin{tabular}{l|ccc}
\textbf{Metric} & \textbf{Clinical} & \textbf{Chemical} & \textbf{Biomedical} \\ \hline
FactScore & .78 & .42 & .42 \\
BARTScore & .35 & .16 & .45 \\
BERTScore-Src & .52 & .47 & .60 \\
\end{tabular}
\caption{Correlation of faithfulness metrics to extractive density of summaries. Correlations computed on the test set of the PRIMERA models after fine-tuning.} \label{tab:extractive-is-faithful}
\end{table}

\begin{figure*}[t]
\centering
\includegraphics[width=\linewidth]{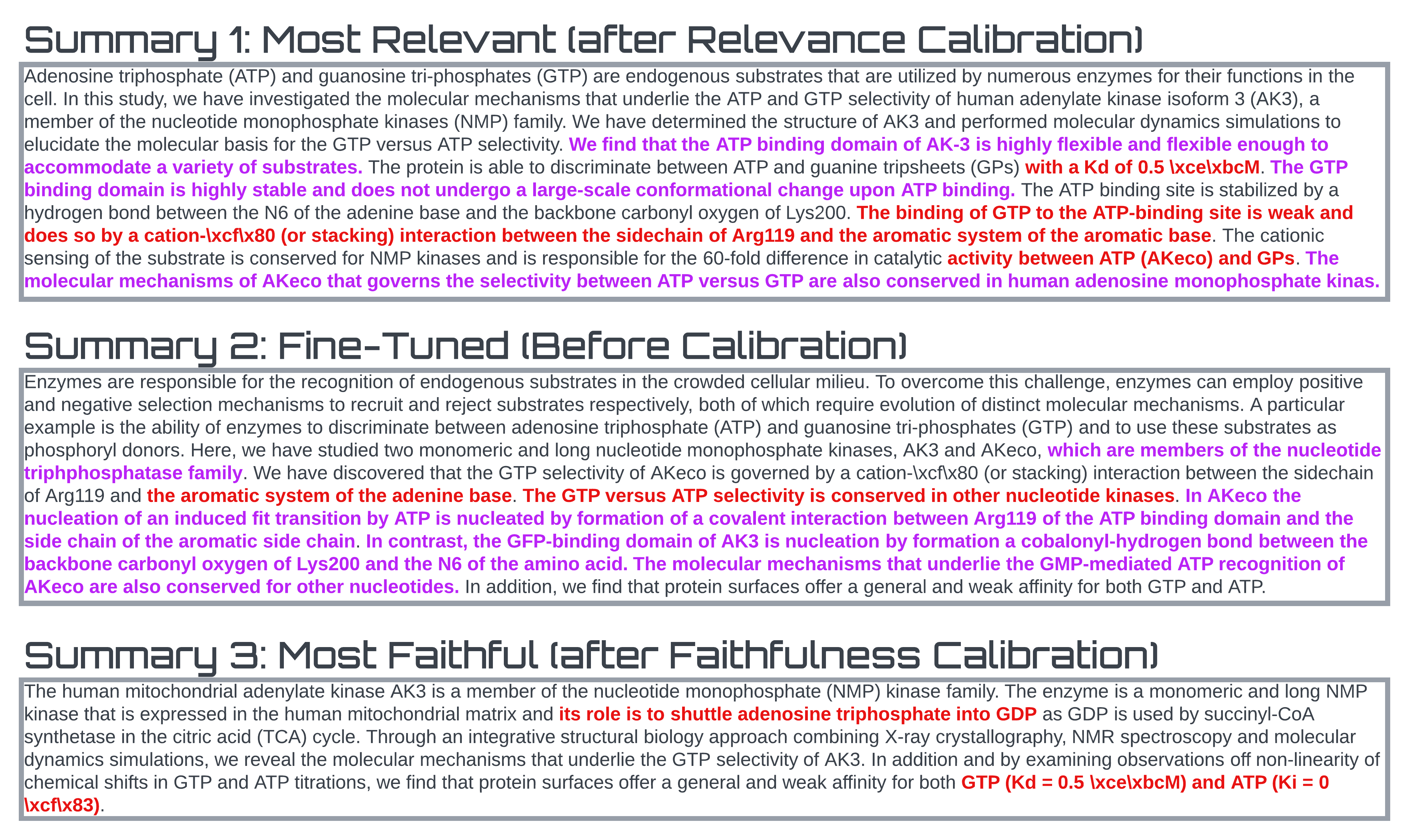}
\caption{Three abstracts generated from model checkpoints after Relevance Calibration (Summary 1), Fine-Tuning (PRIMERA FT checkpoint, Summary 2), and after Faithfulness Calibration (Summary 3). \textcolor{red}{Red Text} has been annotated as being part of an intrinsic error while \textcolor{purple}{Purple Text} is extrinsic.  The annotator rated Summary 1 as the most relevant and Summary 3 the least relevant. } 
\label{fig:human-example}
\end{figure*}

\subsection{Faithful or More Extractive?} \label{app:faithful}

\paragraph{tl;dr.} One would expect that training on contrast sets with a large difference in extractiveness (extractive positives, abstractive negatives) would lead to higher downstream faithfulness. Yet, we find the opposite to be true, which we connect to \S \ref{sec:surprise}.

\paragraph{Evidence.} \citet{ladhak-etal-2022-faithful} note a spurious correlation between the extractiveness of summaries and faithfulness metrics, which holds true for the metrics which make up $Faith_{Agg}$ (as shown in Table \ref{tab:extractive-is-faithful}). One would expect that reinforcing this correlation via contrastive learning (by targeting extractive positives and abstractive negatives) would lead to improved faithfulness metrics. Yet, this does not appear to be the case. Table \ref{tab:faithfulness-results} (\texttt{Spurious} selection type) shows that on average, controlling for a large extractiveness gap does not improve faithfulness ($.131$ versus an overall average improvement of $.133$). If anything, it leads to increased relevance ($.017$ versus $-.067$). While not definitive, a possible driver for this relationship relates to the analysis in \S \ref{sec:surprise}, for which we show that a low likelihood gap between positives and negatives is preferable (an adversarial setup). Since extractive summaries are more likely to be generated than abstractive ones (see Extractive density for Diverse Beam search in Table \ref{tab:method-metrics}), extractive negatives might be preferable to abstractive ones.

\paragraph{Implications.} Given the extractiveness of long-form scientific summaries, more research should focus on subtle faithfulness errors, i.e., those which are less correlated to extractiveness. \citet{zhang2022extractive} provide a helpful typology of errors in fully extractive systems, which can provide a blueprint for the design of more extractive synthetic errors.

\section{Human Evaluation Details} \label{app:human}

To better understand whether or not our calibration models are driving meaningful changes in quality, we conduct a human evaluation on the chemistry dataset. Specifically, we randomly select 50 papers from the test set and collect model generated abstracts from the \texttt{FT} checkpoint as well as most relevant (\texttt{Random} strategy) and most faithful (\texttt{Hard} strategy) \texttt{CT} weights. After randomly shuffling the order of abstracts, we ask each annotator (four authors of this paper with PhDs in chemistry-related fields) to first read the main paper and then, separately for each paper, highlight spans of abstracts containing errors (intrinsic or extrinsic), before ranking the summaries by Relevance \citep{fabbri-etal-2021-summeval}. We defined relevance as in SummEval: \textit{how well does the summary captures the key points of the paper? Consider whether all and only the important aspects are contained in the summary.}. We collect fine-grained faithfulness annotations, rather than summary-level, due to the length of the summaries and prior work on inter-annotator agreement scores of fine-grained errors \citep{pagnoni-etal-2021-understanding, goyal-durrett-2021-annotating}.

\subsection{Error Analysis} \label{sec:error-analysis}

In this section, we analyze the errors from an example in the human annotation set. The abstracts are shown in Figure \ref{fig:human-example}.

Abstract 1 takes the general form of an abstract, providing a reasonable motivation for the work then listing a number of key findings. It makes a number of errors in stating the key findings, however. First, the model seems to have had difficulty with abbreviations and measured values, misreporting a binding constant and confusing GTP and ATP on several occasions. Finally, the model includes several statements not supported in the text. Abstract 2 contains superior prose to Abstract 1, better enumerating the motivation for the work and providing a cleaner concluding statement. It suffers from similar shortcomings, however, confusing GTP and ATP on several occasions and making a number of unsupported claims. In some cases, the unsupported claims appear lifted whole-cloth from another publication. In total, we judge the errors in Abstract 2 to be more misleading than those made in Abstract 1 and thus find Abstract 1 to be more relevant. Abstract 3 is substantially shorter than either Abstract 1 or Abstract 2, minimizing the absolute number of errors it contains. Like the others, it has difficulty with both abbreviations and measured values, making errors due to both. Overall, Abstract 3 is not terribly written; however, its terseness leaves a highly limited description of the paper's contributions. For this reason, it is less relevant than either Abstract 1 or Abstract 2.

\section{Connecting Metric Margins to Diversity} \label{sec:connections}

Larger margin gaps are related to diversity as lexically similar summaries will have similar metric values. In fact, we can examine the \texttt{Diversity} section of Table~\ref{tab:relevance-results} and note that average $Rel_{Agg}$ score across datasets is higher when lexical diversity is maximized ($.114$) than when it is minimized ($.082$). Yet, this trend only holds for the Chemical dataset. To get a more complete sense, we examine the impact of set diversity across runs and note a slightly more reassuring trend: a pearson correlation coefficient of $.21$, $.51$, and $.1$ for clinical, chemical, and biomedical. Interestingly, chemical has the strongest positive relationship between diversity and downstream relevance across runs, yet is negative when directly controlling for diversity.

\end{document}